# Multi-center End-to-End Medical Data Mining with FL-Net - A One-stop Shop for Federated Learning

Simon Süwer[1*], Julian Klemm[1*], Elisa Acitelli[2], Mathieu Almeida[3], Lucia Altucci[4], Zsolt Bagyura[5], Michelangela Barbieri[6], Zsolt-Zoltán Bedő[7], Rosaria Benedetti[4], Béla Bihari[7], Csongor Csalóka[7], Lucia Dicunta[1], Stanislav Ehrlich[8], Bjoern M. Eskofier[9,10], Sándor-József Fejér[7], Georg Fröwis[11], Walter Hötzendorfer[11], Alexandra Kautzky-Willer[12], Jens Johann Georg Lohmann[1], Marianna Maranghi[2], Lorenzo Marconi[13], Rudolf Mayer[14], Wouter Leonard Megchelenbrink[4], Monika Moga[7], Adham Mottalib[15], Sanjeev Mehta[15], Madeleine Müller[11], Thomas Nyström[16,17], Balázs-Attila Orbán[7], Micheal O'Riordain[18], Paul O'Toole[19], Giuseppe Paolisso[6], Paolo Parini[20,21], Matteo Pedrelli[20,21], Enrico Petrillo[22], Philipp Poindl[11], Niklas Probul[1], Anastasia Pustozerova[14], Tanja Šarčević[14], Liane Tinsley[15], Lukas Weilguny[3], Jan Baumbach[1,23†], Andreas Maier[1†]

[*] The first two authors should be regarded as Joint First Authors.
[†] The last two authors should be regarded as Joint Last Authors.

[1]Institute for Computational Systems Biomedicine, University of Hamburg, Hamburg, Germany
[2]Department of Translational and Precision Medicine, Sapienza University of Rome, Italy
[3]Université Paris-Saclay, INRAE, MGP, 78350, Jouy-en-Josas, France.
[4]Department of Precision Medicine, University of Campania Luigi Vanvitelli, Naples, Italy
[5]Institute for Clinical Data Management, Semmelweis University, Budapest, Hungary
[6]Department of Advanced Medical and Surgical Sciences, University of Campania Luigi Vanvitelli, Naples, Italy
[7]SC Egnosis SRL, Sfântu Gheorghe, Romania
[8]Department of Clinical and Movement Neurosciences, UCL Queen Square Institute of Neurology, London, UK
[9]Chair of AI-supported Therapy Decisions, LMU München, Munich, Germany
[10]Institute of AI in Medicine, LMU Hospital, Munich, Germany
[11]Research Institute – Digital Human Rights Center, Vienna, Austria
[12]Division of Endocrinology and Metabolism, Department of Medicine III, Medical University of Vienna, Vienna, Austria
[13]Department of Computer, Control and Management Engineering, Sapienza University of Rome, Rome, Italy
[14]SBA Research gGmbH, Vienna, Austria
[15]Clinical, Behavioral and Outcomes Research Section, Joslin Diabetes Center, Boston, Massachusetts, USA
[16]Department of clinical science and education, Karolinska Institutet, Stockholm, Sweden
[17]Department of Internal Medicine, Södersjukhuset, Stockholm, Sweden
[18]Department of Surgery, Mercy University Hospital and University College Cork, Cork, Ireland
[19]School of Microbiology and APC Microbiome Ireland, University College Cork, Cork, Ireland
[20]Medicine Unit of Endocrinology, Theme Inflammation and Ageing, Karolinska University Hospital, Stockholm, Sweden
[21]Cardio Metabolic Unit, Department of Laboratory Medicine, and Department of Medicine at Huddinge, Karolinska Institutet, Stockholm, Sweden
[22]Mass General Brigham, Harvard Medical School, Boston, Massachusetts, USA
[23]Institute for Mathematics and Computer Science, University of Southern Denmark, Odense, Denmark

# Abstract

Federated learning (FL) enables collaborative training without sharing sensitive data, yet most studies remain simulations. We derived five requirements for successful real-world FL from existing literature, analyzed 14 frameworks, and found that none fully satisfies them.
We present FL-Net, a novel federated framework fulfilling all those requirements. It integrates modular data harmonization, privacy preserving data discovery, disclosure control, securely built, versioned, reusable *Tools,* and containerized *Workflow* execution into a persistent, sustainable network.
FL-Net's end-to-end capabilities were evaluated through harmonization, cross-study patient discovery across MIMIC and US-130, and reproducible, audited federated *Workflows* with up to 50 participating *Sites*. FL-Net is available for self deployment (https://github.com/FedLearnNet) or can be used via its maintained instance (https://federated-learning.net/).
FL-Net will be deployed and live tested across 10 hospitals from 9 countries and used within two European consortia as a practical foundation for interoperable, reproducible, and privacy-preserving multicenter clinical research.

# Introduction

The development of robust machine learning (ML) models necessitates the availability of sufficiently large and heterogeneous datasets, to capture the variation present across patient populations, clinical practices, and healthcare systems. However, clinical data is distributed, sensitive, and tightly regulated to adequately protect everyone's foundational right to privacy (*1*, *2*). Federated learning (FL) offers a conceptually elegant solution to this dilemma (*3*).
FL allows multiple institutions to train a shared model collaboratively without the need to combine their patient-level data. To achieve this, each institution trains the model locally on their *Site* (for definitions of all italicized nomenclature introduced in this work, refer to the Information Glossary, S6). Each *Site* only exchanges model parameters, which are then aggregated to improve the shared model.

However, FL is barely applied in real-world studies. This is evident from various systematic reviews. One review, which examined more than 20 thousand data records and included 612 studies, found that only 5.2% of the studies related to clinical applications in practice (*2*). Another study found that only 23.1% of the 13 evaluated studies adequately described patient inclusion criteria, while limited access to code and data hindered external validation (*1*). Furthermore, FL studies in healthcare often lack *Site* authentication and a fully specified implementation of institutional governance, thereby limiting reproducibility, institutional trust, and real-world deployment (*4*). Overall, based on these findings, none of the existing FL frameworks currently meets all the demands of federated clinical research.

Based on existing literature, we define the main requirements to federated clinical research frameworks: **(RE1)** Heterogeneous *Site* information systems require semantic and syntactic harmonization as well as systematic quality control (*5*, *6*); **(RE2)** *Site* updates, metadata, and outputs require auditable privacy and security controls (*2*, *4*, *7*); **(RE3)** federated workflows should remain reusable and versioned across *Sites* and models (*8*, *9*); **(RE4)** suitable data feasibility and subsetting must be established before training without exposing patient records (*2*, *8*); and **(RE5)** approvals, versions, participating *Sites* and resulting models must remain traceable to support reproducible and accountable studies *(7, 8)*.

After defining clear criteria for the partial or complete fulfillment of these requirements and analyzing 14 FL frameworks (Table 1), we found that no existing framework fully satisfies all requirements. The evaluation criteria and the detailed analysis of each framework are provided in S5. We therefore introduce FL-Net (fig. 1), a novel FL framework to fulfill all previously defined requirements and built with and for two large EU project consortia (fig. 1) (*10*, *11*).

**Table 1 - Feature comparison of representative FL frameworks:** ✓ indicates the requirement is fully implemented, *(✓)* indicates that the capability exists, but it is limited or requires substantial external work, – indicates the requirement is not fulfilled. Cell entries reflect the authors' interpretation of each framework's published description as described in the analysis provided in S5, which also describes the criteria used to determine full, partial or no fulfillment of the requirements. *No longer maintained (<10 commits on GitHub in the last 2 years). †Closed-source software.

| **Framework** | **RE1** | **RE2** | **RE3** | **RE4** | **RE5** |
|---|---|---|---|---|---|
| †Apheris (*12*) | – | ✓ | *(✓)* | *(✓)* | ✓ |
| †CAFEIN (*13*) | – | *(✓)* | – | – | ✓ |
| DataSHIELD (*14*) | *(✓)* | *(✓)* | *(✓)* | *(✓)* | ✓ |
| *FATE (*15*) | – | *(✓)* | *(✓)* | *(✓)* | *(✓)* |
| FeatureCloud (*16*) | – | *(✓)* | ✓ | – | *(✓)* |
| Fed-BioMed (*17*) | *(✓)* | ✓ | *(✓)* | *(✓)* | ✓ |
| Flower (*18*) | – | *(✓)* | *(✓)* | *(✓)* | ✓ |
| MedPerf (*19*) | *(✓)* | – | – | – | – |
| NVIDIA FLARE (*20*) | *(✓)* | ✓ | *(✓)* | *(✓)* | *(✓)* |
| OpenFL (*21*) | – | *(✓)* | *(✓)* | *(✓)* | *(✓)* |
| *Personal Health Train (22) | – | ✓ | ✓ | – | *(✓)* |
| PySyft (*23*) | – | ✓ | *(✓)* | *(✓)* | *(✓)* |
| †Rhino FCP (*24*) | *(✓)* | ✓ | *(✓)* | *(✓)* | *(✓)* |
| Vantage6 (*25*) | – | ✓ | *(✓)* | *(✓)* | ✓ |
| **FL-Net** | ✓ | ✓ | ✓ | ✓ | ✓ |

FL-Net integrates the entire lifecycle of FL into a persistent, star-shaped network infrastructure (FL-Net's implementation details can be found in S4). A central FL-Net *Platform* coordinates all steps, while the *Sites* retain their data and decision-making autonomy at all times. As a result, institutions need to only maintain a single, outgoing, authenticated connection to the *Platform*.

FL-Net describes clinical data semantically and syntactically (S1, Method Figure 6). *Ontology-Nodes* ($N_O$) define clinical concepts and their relationships (*5*, *26*), while *DataType-Nodes* ($N_{DT}$) specify permissible values and validation rules. A *Schema* is organized as a tree of *Schema-Nodes* ($N_S$) that either groups subordinate nodes $N_S$ or terminates in a leaf defined by a $N_O$-$N_{DT}$ pair called $P_{O\text{-}DT}$ (fig. 3A, Method fig. 6A).
Since the architecture is ontology-agnostic, established or organization-specific ontologies can be used (*27*, *28*).

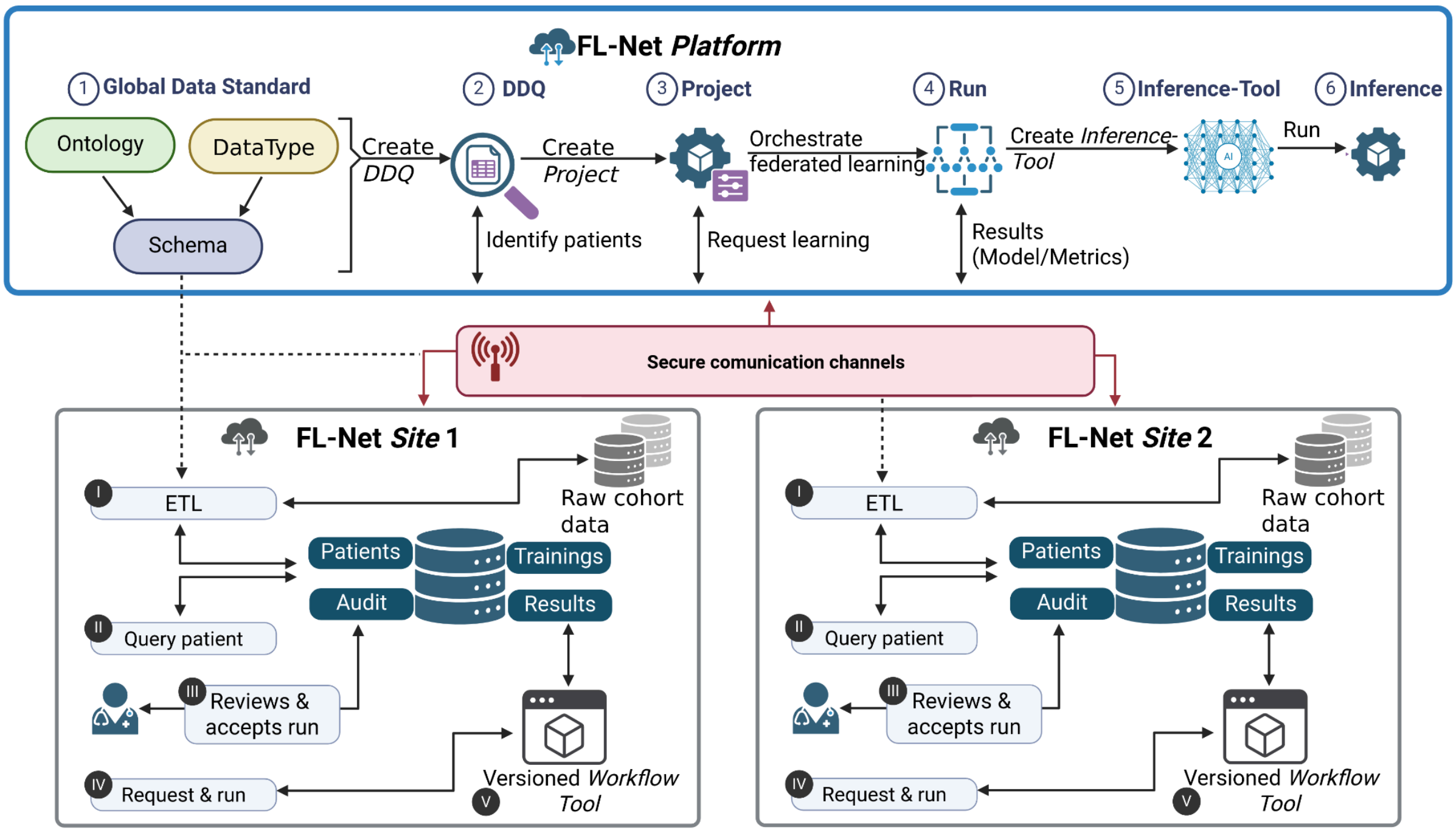

**Figure 1 - FL-Net for federated learning:** The *Platform* combines (1) a globally defined data standard with decentralized institutional *Sites*. *Ontology-Nodes* ($N_O$; green), *DataType-Nodes* ($N_{DT}$ ;yellow), and *Schemas* ($N_S$; blue) form the basis for distributed data harmonization, via extract-transform-load (ETL) and the identification of suitable patient data. At the *Platform*, patient data availability is queried via data discovery queries (*DDQs*) (2), *Projects* are configured, containing the connected *DDQ*, selected variables, and *Workflow* (3), and FL runs are orchestrated (4). The corresponding requests are transmitted to the participating FL-Net *Sites* via secure communication channels. There, the clinical data is organized in cohorts and remains under *Site* control: cohorts are created and data is imported locally (I), patient data discovery queries are executed returning only aggregated counts (II), requests are verified (III), learning is started (IV), and *Workflow Tools* are launched in an isolated execution environment (V). *Site-specific* training and analysis results are logged and returned to the *Platform* exclusively as derived results, model parameters, or metrics and only when publication of results to the Platform was agreed to. An aggregated model is then versioned and stored in the central artifact repository of the *Platform* (5). Audit information accompanies the processing steps and enables traceability of data access, approvals, and executions. The stored *Inference-Tool* can be used by other users and executed on the *Platform* with user-provided data, providing an easy, no-code entry point (6). The *Inference-Tool* stores information of the connected *Project* to facilitate reproduction.

Local, *Site-specific* clinical data goes through a modular, reproducible, validated and auditable extract, transform, load (ETL) process with an added step of mapping columns to $N_S$. Invalid values are rejected and reported back to the responsible users based on validation rules of the associated $N_{DT}$. Subsequently, the transformed, annotated, and validated data is stored locally in a longitudinal entity-attribute-value (EAV) database that links $N_S$, patients, visits, time points, and connectors. Patient data is organized into one or multiple cohorts at each *Site*. This allows institutions to maintain different source systems while still providing a common internal representation for all processes.

The FL process begins with the semantically controlled selection of suitable data (fig. 2). *Researchers* formulate an ontology-based data discovery query (*DDQ*) whose elements refer to global $P_{O\text{-}DT}$ (S2). Each *Site* resolves these references against its own *Schema* and uses them to generate a local SQL query across its EAV database. The logical *DDQ* thus remains unchanged across different institutions and local data structures. Before counts are returned to the *Platform*, the *Site* applies privacy and security measures to protect from identification. The *Site* checks cohort-specific permissions and enforces minimum thresholds for patient counts, with small patient counts suppressed and otherwise rounded (*29*, *30*). *DDQs* and matching patients are logged locally for auditing. The *Platform* receives only controlled, aggregated counts or separately requested statistics. The Site specific statistics are protected by the same security mechanisms used by the *DDQ*. These show whether institutions can contribute suitable data, are potentially biased, and which *Schema* elements are available for the planned learning without revealing patient-level information.

If a sufficient number of patients are discovered, the *DDQ* can be the starting point for a federated *Project* (fig. 2). Here, *Researchers* can define a set of variables to use for FL based on the *DDQ* related *Schema* information. The *Project* is then linked to a versioned *Workflow*. A *Workflow* is a directed acyclic graph of *Tools* (*31*). Each *Tool* is a versioned, reusable, and auditable application with defined inputs and outputs, hyperparameters, publication, and auditing statuses (*31*). *Tools* can be used for extraction or transformation of patient data inside each *Sites* ETL process. *Tools* may also serve as *FL-Tools* achieving preprocessing, analysis, ML, and other data analysis tasks. Lastly, trained models may be used to create *Inference-Tools* if *Sites* agreed in the approval stage, which can either be used in FL or by users directly (*31*). *Researchers* can select suitable *Tools* without having to create a new execution environment for each *Project*.

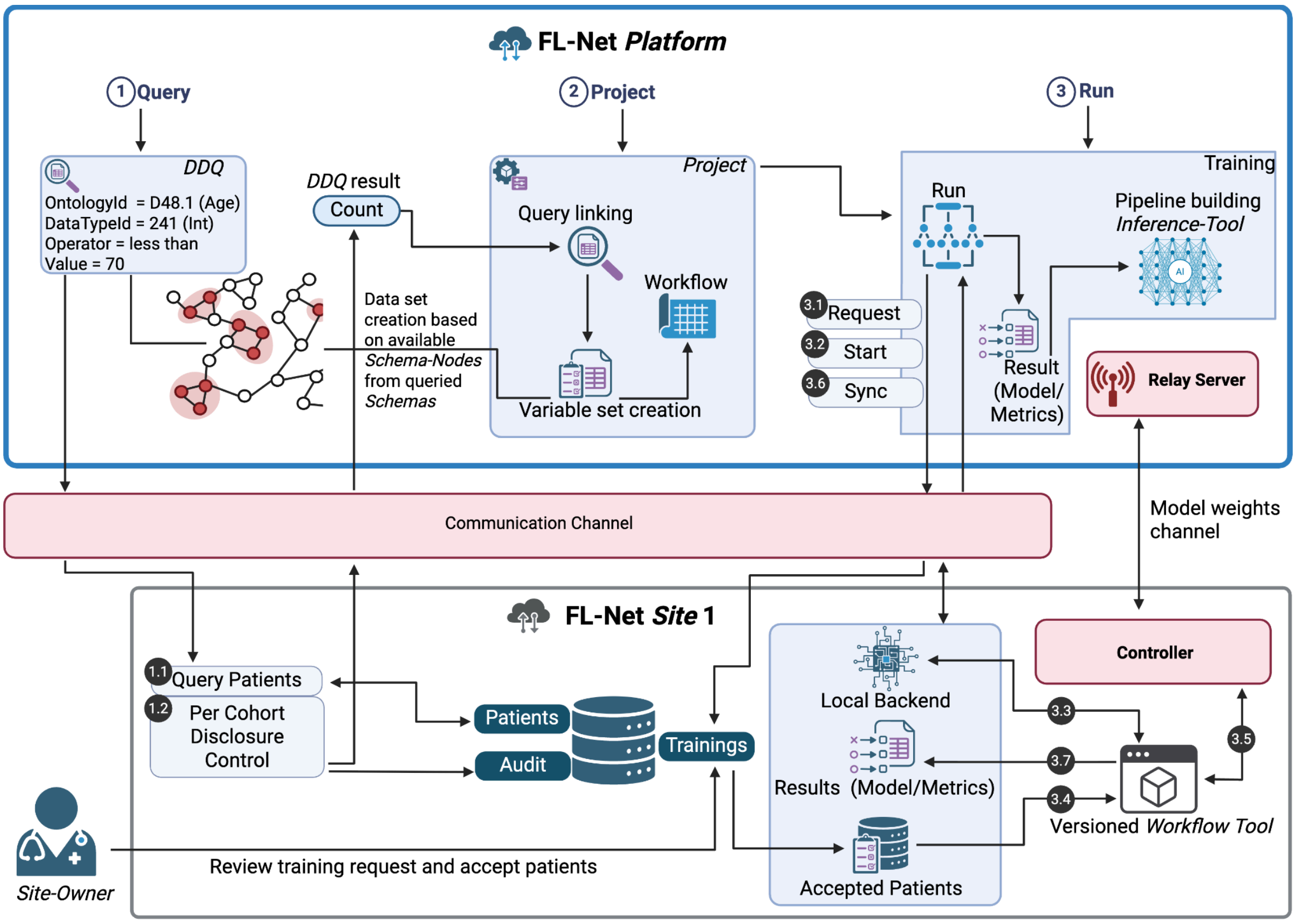

**Figure 2 - Federated Learning Workflow and Execution Architecture in FL-Net:** The FL process is initiated by sending an ontology-based *DDQ* to select semantically and syntactically defined subsets of cohorts (1). The *DDQ* is linked to a *Project*, in which an eligible variable set can be created from *DDQ*-related *Schemas* and used with a versioned federated *Workflow* (2). Subsequently, the FL-Net *Platform* creates and coordinates an FL *Project* run (3). Initially, an *FL-Request* is sent. *Sites* are responsible for independently reviewing and authorizing *FL-Requests* and approving patients for the FL *Project* run. At any time patient-level data remains within institutional boundaries. Counts of approved patient data for the FL *Project* run are sent to the *Platform*, undergoing the same disclosure control as the *DDQ*. Once sufficient *Sites* have agreed (default 3) and sufficient patients are available, the *Researcher* can start the FL *Project* run. Running the FL *Project* locks the set of agreed *Sites*, other *Sites* agreeing later are left out. During the FL *Project* run, local model updates are sent to the coordinator, aggregated, and redistributed for subsequent rounds. *Tool* execution is secured by an auditing system as well as being executed in isolation with limited network access. All *Workflows*, *Tools*, runs, metrics, and resulting models are versioned and persisted, thereby ensuring reproducibility, traceability, and governance throughout the learning lifecycle. The resulting models and their metrics may only be available at participating *Sites* depending on the agreed-upon *FL-Request* for privacy and IP-right reasons.

Once a *Project* is configured, the *Researcher* can send an *FL-Request* (fig. 2). The participating *Sites* review this request independently and decide whether and to what extent they will participate. Training can be started by the *Researcher* only after a sufficient number of *Sites* and thereby patients have been approved. The acceptance count is subjected to the *DDQ* disclosure control mechanism. *Tool* execution always happens in isolated containers and accesses only the approved patients and defined networks. Patient data never leaves the *Site*. The *FL-Request* specifies whether the resulting model is published or stays only at the participating *Sites.*

The results of the run are then returned to the participating *Sites* and saved as a local artifact. In addition to the model metrics, *Project* information, run statuses, and cohort provenance are stored.

FL-Net makes it possible to use heterogeneous clinical data from multiple *Sites* for FL. Each dataset has been imported via quality-assured ETL processes. For suitable patient data discovery, FL-Net uses the *DDQ*. Training facilitates the automatic documentation of *Workflows*, approvals, versions, and results, enabling reproducible and responsible research. The resulting models can be directly converted into *Inference-Tools* within the *Platform* and made available to other *Researchers* for inference processes. Thus, FL-Net harmonizes data across local *Sites*, supports *Researchers* in data retrieval, simplifies and isolates the training process, and manages FL training results and reproducibility parameters via the *Tool-Store*.

# Results

## FL-Net implements a hierarchical Schema system for semantic and syntactic data harmonization (RE1)

To enable interoperable data harmonization across heterogeneous partner *Sites*, we developed a semantic and syntactic framework for FL-Net, inspired by BioCypher (*32*). Each cohort is linked to a *Schema* (fig. 3A, Method fig. 6A).
Local raw data is harmonized towards the target *Schema* through no-code, reusable, and audited connectors that define the ETL steps (fig. 3B). Site-Owner can optionally select transformations as needed, must map columns to $N_S$, and optionally attach metadata, such as timestamps. During import, the data is transformed and validated against $N_S$ constraints. Invalid values are rejected and reported. *Sites* that use the same *Schema* are harmonized by design, while different *Schemas* are partially harmonized wherever the same $P_{O\text{-}DT}$s are used.
We evaluated FL-Net's semantic and syntactic harmonization using the inherently different US-130 diabetes (*33*) and MIMIC-IV 2.2 ICU datasets (S1) (*34–36*). We preprocessed US-130 subsets in different ways to demonstrate the varying capacities and adaptability of the ETL process. This adjusted version is referred to as “modified US-130” from hereon.

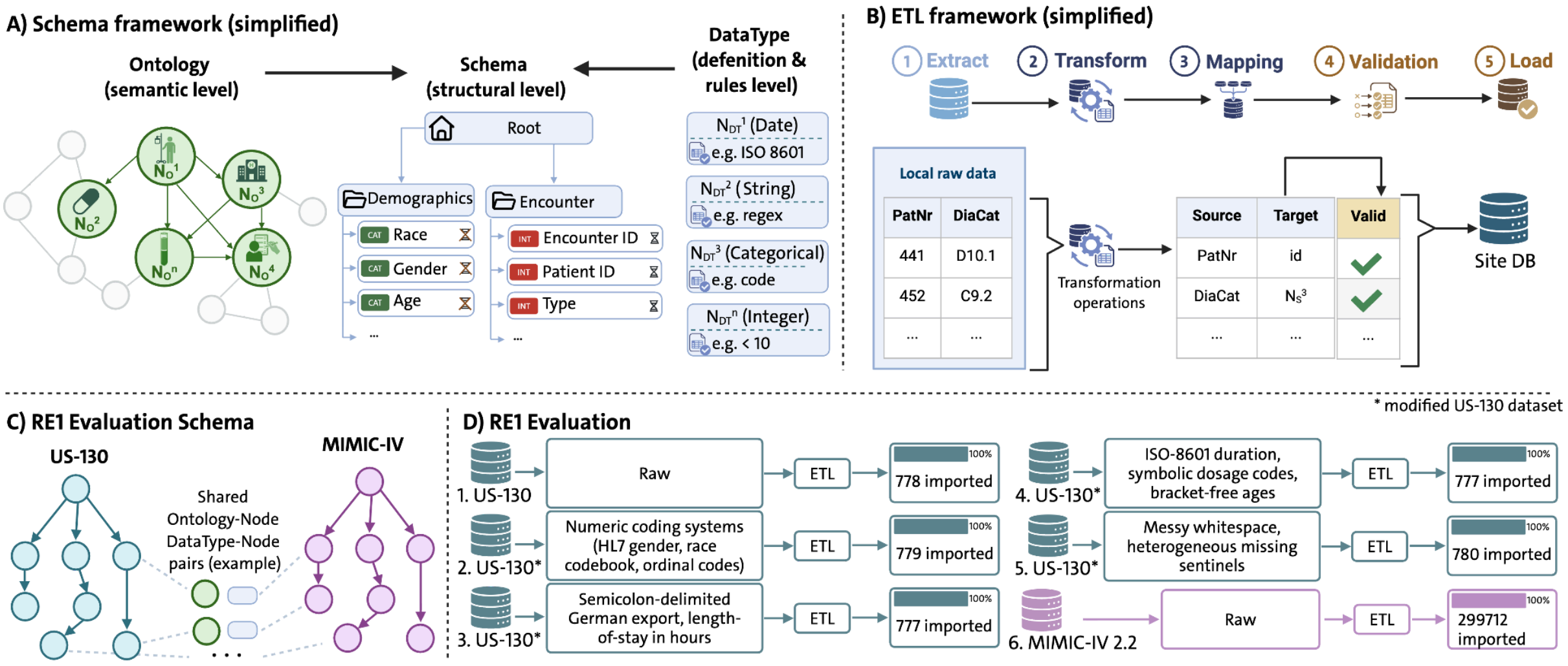

**Figure 3 - Semantic and syntactic harmonization and its evaluation across heterogeneous datasets:** A) FL-Net represents data *Schemas* through linked $N_O$ and $N_{DT}$ described by an $N_S$ (Method fig. 6). B) Local source data is harmonized through reusable ETL connectors that extract, transform, validate, and load data into the target *Schema* (Method fig. 6). C) Structurally different *Schemas* remain interoperable when their $N_S$ resolve to shared $P_{O\text{-}DT}$, enabling equivalent $N_S$ to be identified across datasets and standards. The *Schema* used for US-130 and MIMIC-IV datasets are 100% overlapping. D) Harmonization was tested on the US-130 diabetes dataset, modified US-130 variants representing heterogeneous encodings and source formats, and the MIMIC-IV v2.2 ICU dataset. Details can be found in S1. All of the evaluated data were successfully transformed into *Schema-compliant* representations. Despite differences in coding systems, formatting, and missing-value conventions, 100% of the evaluated records were imported.

All *Sites* achieved complete, *Schema-compliant* representations through configuration alone (fig. 3 C-D, Table S1.3), successfully mapping all required variables. Importantly, harmonization was not restricted to a single shared *Schema*. A 100% overlap of the MIMIC-IV and US-130 dataset *Schemas* could be achieved, making the distinct US-130 and MIMIC-IV datasets completely interoperable within FL-Net purely by configuration of ETL connectors, transforming source data, and mapping to the unifying *Schema*. Consequently, validation against type constraints and allowed value domains yielded no rejections or type errors. This demonstrates that complex and differently structured cohorts, such as US-130 and MIMIC-IV 2.2, can be represented within the same harmonization framework.

## Layered privacy and security, including auditing with fine grained Site-Owner controls (RE2)

To ensure that *Site-Owners* retain full control over their local patient data, FL-Net provides *Site-controlled* disclosure safeguards with any local interaction from *DDQ* to *Project* execution safeguarded. *Site-Owners* define for each cohort which *Researcher* may search the contained patient data. Unauthorized *DDQs* are rejected. To prevent re-identification attacks that exploit small, deliberate changes to *DDQ* parameters, *Site-Owners* additionally set a suppression threshold (default 100). Results below this suppression threshold are reported as zero. Remaining positive counts (c) are rounded up to the nearest multiple m(c) of the rounding step, defined as:

$$m(c) = 10^{\lfloor \log_{10}(c) \rfloor - 1}$$

The reported privatized count p(c) is calculated as:

$$p(c) = m(c) \lceil \frac{c}{m(c)} \rceil$$

For example, 123 becomes 130 and 5555 becomes 5600; exact multiples remain unchanged (S2.2). Furthermore, the *Site-Owner* can configure a minimum response time interval for repeated queries (see Methods - Security and Privacy). The same cohort-level access permissions are checked when evaluating an *FL-Request* to determine whether it may use part of the cohort for FL. Such an *FL-Request*, which contains the *Workflow,* can only contain any valid combination of *Tools*. At accepting the request, the *Site-Owners* can additionally exclude individual patients from this specific FL run. Every *Tool* is built through the FL-Net Building-Pipeline, undergoes automated vulnerability/malware scanning, and can be audited by *Auditors*. During execution, *Tools* run in a containerized, encapsulated environment and can only communicate with authorized components and networks, with no internet access by default (*31*). All FL communication of local models towards other *Sites* or the aggregator is transport and machine to machine encrypted (see Methods - Security and Privacy). Each interaction with patient data is logged. This includes changes to patient data within the *Site*, as well as interactions initiated through the *Platform*, such as *DDQ*, and FL *Projects* that are requested or executed, together with the *Tools* and versions involved. *Site-Owners* can search this record at any time providing a persistent, verifiable log of how and by whom the data they govern was indirectly used.

## Suitable patient discovery mechanisms must be available before training without exposing patient records (RE4)

To enable federation-wide data exploration, FL-Net provides a user-centered discovery system for *cross-Schema* and *cross-Site DDQs*. Each *DDQ* contains multiple conditions. Each condition targets a specific $P_{O\text{-}DT}$ and supports logical or mathematical operators with optional reference values. Conditions are combined using logical conjunction to identify matching patients and return aggregated counts.
Individual patient records remain on the Sites and the previously described disclosure controls protect against re-identification.

To evaluate the *DDQ*, we sent several queries to four test *Sites* ( S2.3). Three *Sites* included US-130 patients, and one included MIMIC-IV patients. Each *Site* resolved and executed the *DDQ* locally, counted matching patients, applied its suppression-and-rounding policy, and returned only the approved aggregate counts correctly.
As can be seen in the Table S2.1, the repeated runs yielded identical results, which confirms the deterministic *DDQ*. Counts above the configured threshold were rounded up, while small counts were suppressed and reported as zero. Since the same *DDQ* was sent to different *Sites* with different *Schemas* (but overlapping $P_{O\text{-}DT}$, fig. S1.1) and could be executed by them, we were able to demonstrate the *cross-Schema DDQ* functionality. This allows data to be discovered and FL to be performed even with differently structured cohorts simply by aligning to existing *Schemas*, or overlapping $N_S$ of different Schemas. Consequently, FL-Net enables reproducible semantic patient data discovery across different *Schemas* while preventing identifying counts being shared by the *Sites*.

## Comprehensive versioning, auditing, and artifact management ensure traceability and reproducibility (RE3 & RE5)

FL-Net enables the execution of federated *Workflows* consisting exclusively of verifiable, versioned *Tools* from the *Tool-Store* (*31*). For each training session, the *Site-Owners* are requested to participate and for their decision provided with the immutable Projects description and configuration. The *FL-Request* is only displayed to the *Site-Owner,* if the *Site* has found patients for the linked *DDQ*. Out of all matching patients at learning request arrival, the *Site-Owners* can further adjust which patients should participate in the training. This ensures that only patients approved by the *Site-Owner* are included. The *Site-Owner* can review the patient variables (defined by $N_S$) and which *Tools* are used in the *Workflow*.
Upon completion of a *Project*, execution logs, metrics, and associated artifacts, like the model weights or *Tool* reports, are archived at the respective *Sites*. With the consent of all participating institutions, *Inference-Tools* can be published globally in the *Tool-Store*, linked to *Project* information, and stored with all necessary artifacts produced by the FL. This allows analyses to be reproduced and models to be reused in a targeted manner.

We evaluated the FL *Workflow* system by training MLPClassifier for binary binary 30-day readmissions prediction (*37*). Three training approaches with identical data partitions were compared (fig. 5): I) Isolated single *Site* training; II) federated training; and III) centralized training on pooled data. The exact evaluation setup is shown in S3. To evaluate FL-Net's performance across a varying number of *Sites*, the US-130 clinical dataset was uniformly distributed among 130 synthetic clinics by sequentially assigning equal-sized contiguous subsets of the data to each clinic.

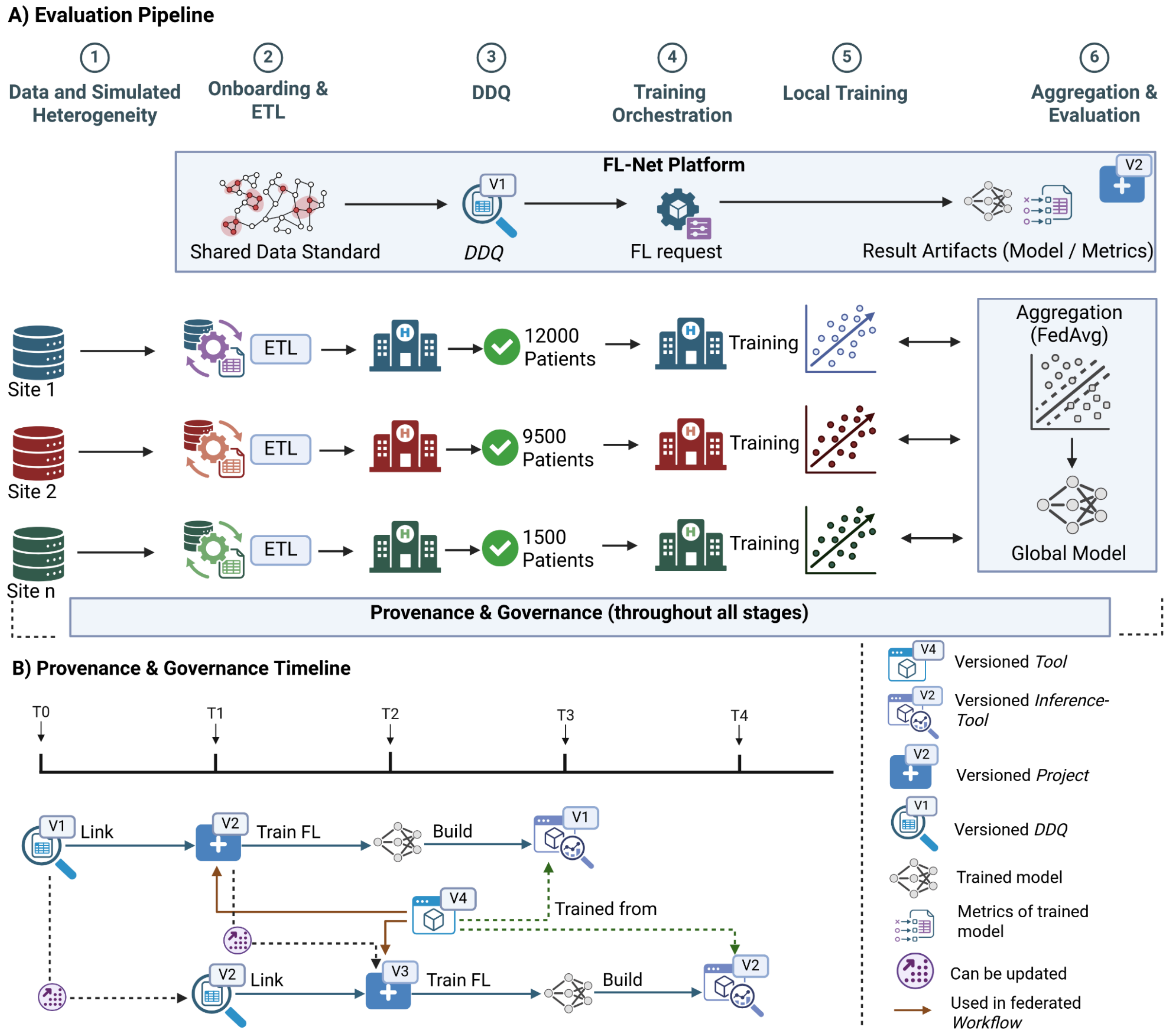

**Figure 4 - Evaluation A) and provenance B) Workflow for FL-Net: A)** The evaluation setup follows six sequential stages: (1) The modified US-130 dataset is distributed across up to 50 simulated clinical *Sites*. The dataset was split by clinic. (2) Each *Site* performs ETL, also mapping local variables to $N_S$, and validates the resulting data against respective $N_{DT}$ definitions. (3) *DDQs* are sent to all *Sites*, executed under disclosure control, and only returned as aggregated patient counts. (4) Based on the aggregated discovered patient numbers, the global *Researcher* defines and sends out a *Project* through an *FL-Request*. This request contains information on the exact data that would be used by which Tools, and whether a resulting model should be public or stay at the *Sites*. (5) Each *Site* trains the model on its own data and sends model updates, such as gradients or weights, to the aggregation layer. For the training set, the data was split using an 80/20 random split. (6) FedAvg is used for federated aggregations. **B)** Changes to *DDQs*, *Projects*, *Tools*, and *Inference-Tools* generate immutable versions, enabling end-to-end traceability. A specific *DDQ* forms the basis of every *Project* in which versioned tools, hyperparameters, and selected semantic data variables are combined into a *Workflow*. After initiating an FL execution, participation and data approvals are obtained from the *Sites*. Upon FL start, the participating *Sites* and approved patient numbers are immutably recorded. Results and artifacts are stored and made accessible in accordance with the granted approvals. A versioned *Inference-Tool* can be built from a completed *Project* run, which retains the complete provenance for reproducible future use. The *FL-Request*, containing the number of approved patients per *Site* (masked with a random ID per *Site*), the public results, and the versioned *Project*, are recorded on the *Platform*. *Sites* store the initial *FL-Request* and all locally created information, including their own local logs and results.

The single *Site* training setting (I) represents institutions operating independently, while the centralized training setting (III) serves as an upper-reference scenario. Federated performance was evaluated based on improvement over isolated training and proximity to the centralized reference. Algorithmic details are provided in S3. At each *Site*, the data were stratified by target class and split into an 80% training set and a 20% internal validation dataset. In the single *Site* training setting (I), only the training data of the specific *Site* was used in training. For the federated (II) and central (III) setting , the 80% training sets from all *Sites* included in the training scenario were combined via federated learning (II) or pooling (III).

Two separate evaluations were conducted. In the first scenario, the models were tested on the internal validation dataset (i) (fig. 5A). In the second scenario, the final federated and central models were tested on an external dataset consisting of 26 unseen clinics (ii) (fig. 5B) hereby called external validation-clinics dataset.

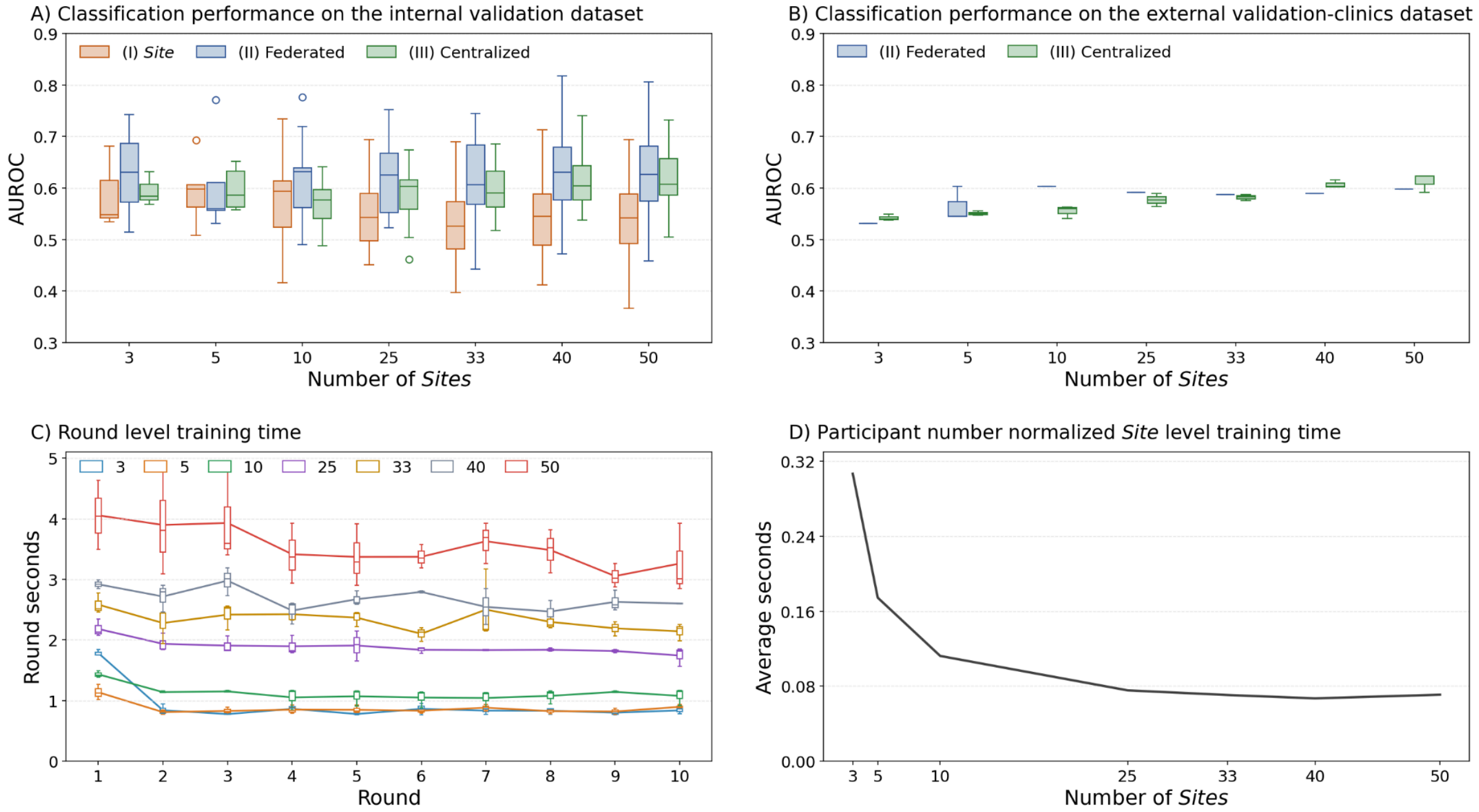

**Figure 5 - Federated evaluation:** A) Shows the AUROC when evaluating on the internal validation dataset (i) given the three training scenarios of individual *Site* (I), federated (II) and centralized (III) on a varying number of *Sites* involved in the training. With increasing number of *Sites, the* performance of the *Site* (I) trained models worsened in comparison to the federated (II) and centralized (III) scenario. B) Shows the AUROC when evaluating on the external validation-clinics dataset using the federated (II) and centralized (III) training scenarios. The mean AUROC ranged from approximately 0.53 to 0.60 for federated models, and from approximately 0.54 to 0.62 for central models. Federated models achieved a higher mean AUROC at 5, 10, 25, and 33 participating *Sites*, whereas central models achieved a higher mean AUROC at 3, 40, and 50 *Sites*. The values shown are averages across three repetitions using 10 training rounds per setting. C) The mean duration of each round generally increased with site size, reaching its maximum with 50 *Sites*. The lines show the mean duration of each of the 10 training rounds across three repetitions, while the boxplots indicate the variability due to repetition. D) The mean round duration, normalized by the number of participating *Sites*, decreased from approximately 0.31 seconds with three clients to around 0.071 seconds with 50 *Sites*. Values in D) were averaged across three repetitions

FL improved classification performance on the internal validation dataset of the participating Sites compared to local training (fig. 5A). When validating on the external validation-clinics dataset, the AUROC gain of the federated scenario increased from +0.041 for three *Sites* to +0.097 for 50 *Sites* as the federation size grew (fig. 5B). Performance varied more widely across the validation clinics. The highest AUROC for the federated model was achieved with 10 *Sites* (0.603; 0.555 for the centralized model), while with 50 *Sites*, the centralized model was slightly superior (0.613; 0.597 for the federated model). As the federated model's performance stays close to the centralized one, we successfully demonstrate that FL-Net is correctly training FL models.

The round-trip time increased with the size of the *Sites* (fig. 5C), from an average of 0.92s with three clients to 3.55s with 50 clients. Thus, a 16.7-fold increase in the number of clients resulted in only a 3.9-fold increase in round-trip time. The round duration per individual round with the same number of *Sites* remained consistent. The average runtime per *Site* per round decreased from 0.31s to 0.076s and plateaued at around 25 *Sites* (fig. 5D). Over the rounds, the runtime remained stable for each *Site* size.

# Discussion

Despite its theoretical viability, the practical deployment of FL remains severely constrained (*2*). Institutions must harmonize heterogeneous data (*1*, *2*, *4*, *6*), identify suitable cohorts (*4*, *6*)), and clarify usage rights (*38*). Rebuilding these processes for each study limits reproducibility and scalability. FL-Net closes this gap by integrating heterogeneous data, enabling semantic data discovery, providing local disclosure control, ensuring isolated execution, and managing resulting artifacts in a single, reproducible, and traceable end-to-end process.
Our evaluation demonstrates this integration in practice (fig. 4). Despite the heterogeneity of the (modified) US-130 and MIMIC-IV datasets, the configurable connectors enabled their conversion into a common, *Schema-validated* representation. Based on this, it was possible to use the same *DDQ* across all *Sites*. Only aggregated patient counts, protected by thresholds, rounding, and rate limiting, were disclosed in this process. The identified patients were then successfully used to train versioned *FL-Tools*. The weights from this training were then transferred to a versioned and linked *Inference-Tool*. The targeted and joint integration of these steps is largely unaddressed in current research (Table 1, S5).

FL-Net is designed as a sustainable research network, not a temporary infrastructure for a single study. Persistent networks can reuse harmonized patient data via their *Schemas* and *Tools* in the *Tool-Store,* reducing integration and coordination costs for subsequent studies. With each additional *Site*, the discoverable data space grows. As the scaling experiments show, FL-Net effectively supports a large number of *Sites* (fig. 5).

Since all *Sites* ran for the validation on a shared server, real-world differences in hardware, bandwidth, security policies, and local administration were not reflected. Additionally, the evaluation is based on a public datasets and a single prediction task. Future work will assess FL-Net in real-world deployment across multi-center scenarios within the dAIbetes and Microb-AI-ome projects.
To streamline privacy preserving data discovery, the automatic resolution of ontology relations for the *DDQ*, e.g. translating queries from supernodes such as cancer to subnodes such as colorectal cancer, should be investigated. Furthermore, a $N_{DT}$ to $N_{DT}$ data translation system should be investigated, making syntactically different data of the same semantic concept automatically interoperable. Finally, future research should explore integrating alternative FL frameworks. Incorporating a computation-focused framework, such as Flower (*18*), would be highly beneficial due to its robust focus on diverse aggregation algorithms and advanced privacy-enhancing techniques.

The next step, therefore, is to establish a real, continuously growing network of *Sites*. Such a network should not only train individual models but also jointly refine cohorts, *Workflows*, and models over time. *Sites* will only participate if the benefits outweigh the integration and operating costs. This includes funded local infrastructure, compensated data curation, access to jointly developed models, and clearly defined rights to and recognition for the resulting artifacts. This is explicitly done in the dAIbetes (*10*) and Microb-AI-ome (*11*) consortia funded by the European Commission HORIZON Europe program. FL-Net is currently being deployed there and will cover 800,000 patients from 10 participating hospitals in nine countries (EU, USA). It is live and tested in two separately hosted networks, demonstrating the on-premises deployability. It is our hypothesis that the open-source nature and comprehensive documentation of FL-Net will serve as a foundation for the future development of sustainable FL initiatives on a global scale, enabling the transition from individual feasibility studies toward an interoperable and institutionally controlled federated research infrastructure.

Finally, FL-Nets harmonization, DDQ and reproducible and modular FL functionalities aim to enforce FAIR principles within FL, given the constraint that FAIR is not directly transferable to FL (*39*). Controlled *DDQs* support findability. Authorized local execution supports protected accessibility. Shared semantic and syntactic representations and API-standardized *Tools* support interoperability. Versioned *DDQ*, *Workflows*, parameters, results, and *Tools* including *Inference-Tools* support reproducibility.

# Methods

## Data standard

A shared data standard substantially reduces the effort of integrating heterogeneous *Sites* (*5*). FL-Net is based on data standards that allow FL-Net to guarantee validated data at *Sites* with full access to the semantic and syntactic data description on the FL-Net *Platform*. FL-Net represents this standard as a tree, hereby called *Schema,* with three coupled node types (Methods fig. 6A). Ontologies define the meaning of clinical concepts and their relationships, establishing the semantic level (*26*). *DataType-Nodes* ($N_{DT}$) define the possible values of a concept and the rules that validate them, establishing the syntactic level (*40*, *41*). A *Schema-Node* ($N_S$) then links an ontology node ($N_O$) to a $N_{DT}$, defining the semantic and syntactic level of a variable. The $N_S$ are organized in a tree-structured *Schema* that specifies which elements belong together and which are expected for a given cohort, establishing the structural level.

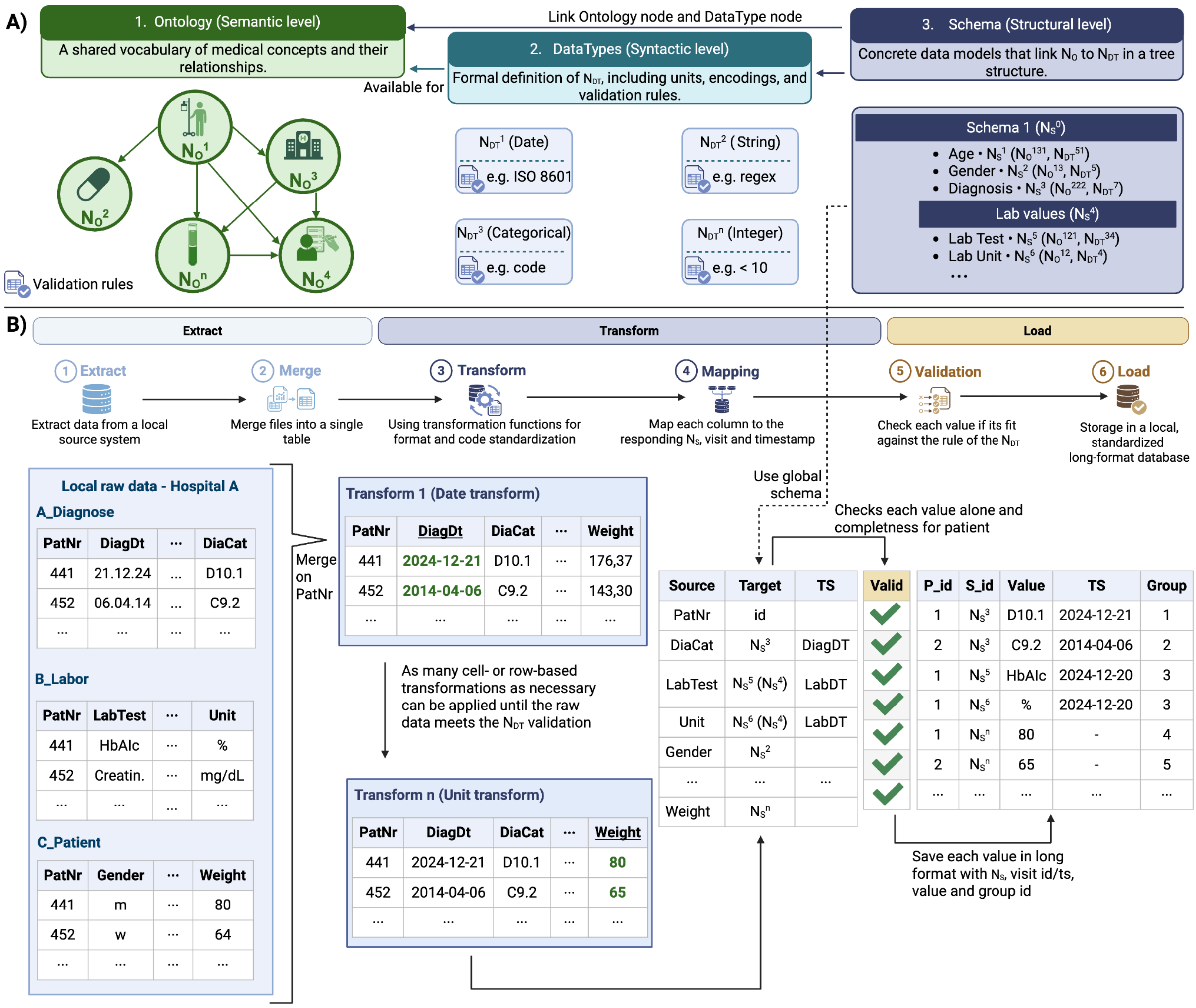

**Figure 6 - Ontology-guided harmonization and ETL Workflow in FL-Net:** (A) Conceptual relationship between the three harmonization layers: (1) $N_O$ define the semantic level by providing a shared vocabulary of clinical concepts and their relationships. (2) $N_{DT}$ define the syntactic level, specifying permissible formats, units, encodings, value ranges, and validation rules. (3) $N_S$ link the semantic and syntactic levels by linking $N_O$ to a concrete $N_{DT}$, defining the structural level. Together, these layers ensure that each *Schema* element is semantically interpretable and technically validatable across *Sites*. B) An example ETL chain for converting a local *Sites* hospital data into the FL-Net target representation: (1) Raw data are extracted from local source systems. (2) Multiple local tables are merged into a single processing table. (3) Transformation functions and *Tools* standardize values, formats, timestamps, codes, and units. Green values show the transformations performed.(4) Source columns are mapped to the corresponding target $N_S$. Additional visit and timestamp information for the target $N_S$ from other columns may be optionally mapped as well. (5) The mapped values are validated against *Schema-linked* $N_{DT}$ rules and checked for patient-level completeness. (6) The validated values are then loaded into a local, standardized, long-format database. Each entry is stored alongside the $N_S$, visit identifier, timestamp, and value. A new P_id is generated per patient and linked to the original patNr given. This internal P_id is used by the system, while the *Site-Owner* is shown the original patNr they provided when interacting with their *Site* software. To avoid losing information from wide format source data row-wise grouping information is stored per entry (ETL description can be found in S1). In the shown example, this would allow to correctly query for a HbA1c lab test with percentage representation.This procedure transforms heterogeneous local data into a common, *Schema-constrained* (semanticly and syntactically validated) representation, enabling federated querying and learning.

FL-Net is ontology-agnostic. While it is possible to use the UMLS Metathesaurus (*42*) as its ontology backbone, it is also possible to work entirely with custom ontologies. This is important because UMLS licensing can create barriers to redistribution and to conducting fully reproducible, openly shareable experiments (*42*).

Each FL-Net *Site* can organize its data in different cohorts. Each cohort follows a *Schema*. Each *Schema* may be compatible with other *Schemas* due to an overlap of $N_O$. This overlap can occur within a single *Site* or across multiple *Sites*. The utilization of global ontology structures administered by the *Platform* facilitates the identification and integration of similar or related cohorts for training purposes, given the established structure. The *Schema* ensures that a *Sites* cohort is complete and consistent, meaning that the expected attributes are valid against their $N_{DT}$ and are actually present. As a result, the data standard does not become a prerequisite outside the system, but rather a technical component of the network itself.

$N_O$ and $N_{DT}$ form the global structure, while the *Schema* organizes them into a tree-based common data standard to which *Site* cohorts can be consistently mapped against. This provides FL with a commonly known and transparent data standard that can be validated across participating Sites (*5*).

## Data harmonization by extract-transfer-load

FL-Net provides extract-transform-load (ETL) functionality to harmonize and load the local data into a *Site*. During the ETL process the *Site* data is converted (transformed) to fulfill the shared data standard (Method fig. 6B).

This is done via so-called connectors. A connector stores information on how raw data should be extracted, which transformations should be applied to which columns, and how the data is mapped post transformation to $N_S$ of a data standard. Both steps, transformation and extraction, can be done using built-in functionalities or (certified) *ETL-Tools* from the FL-Net *Platform* can be used. Each transformation uses either a cell or a row as input, while the extraction uses complete files as input. Both extraction and transformation support *Tool-specific* hyperparameters. In addition, this enables, for example, the existence of an SQL extraction *Tool*. External databases can therefore be connected to FL-Net using specific connectors. By default *Tools* have no internet access, but extraction *Tools* can be launched with specific network access to enable remote data extraction via the *ETL-Tool*.

A connector stores information on extracted, transformed, and final column names, the *Schema* it is created for, and $N_S$ mapping of final column names. This allows error reporting in case an ETL run misses input or produces unexpected intermediary or final results. The connector saves the mapping from data columns onto $N_S$. As data is ingested, every value is checked against the linked $N_{DT}$ validation rules of its target $N_S$. Values that violate the rule are rejected according to the rules and reported back to the user.

Harmonized data is stored locally in a long entity-attribute-value (EAV) format, in which each row records a single observation for a patient, $N_S$, time point, and original row identifier. The original row identifier can in later analysis be used to link previously connected information (such as a diagnosis and the type of diagnosis) ensuring no information loss due to the EAV format. This format is ideal for clinical data as it can accommodate longitudinal, heterogeneous, and sparse data, also incorporating the typical Patient-Visit-Variable hierarchy of Electronic Health Records (EHRs) (*43–45*). The flexibility of this format allows for new variables or repeated measurements to be added without altering the table structure. Institutions with different source data describing the same variables can therefore be mapped to the same common data standard, enabling follow-up data discovery and FL.

## Data discovery query (DDQ)

FL in FL-Net does not begin immediately with training a model, but rather with the semantically and syntactically controlled selection of available patients or samples and characteristics that are required for training.
To this end, FL-Net proposes the use of an ontology-centric query language, an approach used in other contexts already (*46*, *47*). A *DDQ* can be composed by a *Researcher* via the frontend of the FL-Net *Platform*. A *DDQ* is a list of *DDQ* items, where each item represents a condition (DDQ experiments can be found in S2). Each *DDQ* item references a globally available $P_{O\text{-}DT}$. Multiple *DDQ* items can be combined into one *DDQ*. *Sites* resolve these references to their respective *Schema* structure and compile standalone SQL queries against the local EAV-format data store. In this process, cohort-specific permissions are checked before results are returned, minimum patient count thresholds per cohort are enforced locally, and the reported counts are additionally rounded and rate limited. Any *DDQ* is logged locally for auditing on affected patient-level.
*DDQs* allow ontology-based data discovery even across multiple different *Schemas*. Based on *DDQ* results, FL-Net allows *Researchers* to define the set of $P_{O\text{-}DT}$ they want to use as variables in the following FL execution. *Researchers* can only select $P_{O\text{-}DT}$ that are available in the *Schemas* found by their *DDQ*.

## Training

The federated training framework combines containerized *FL-Tools* with versioned *Inference-Tools* (fig. 4). *Workflows* and *Tools* are managed as standalone *Platform* objects with defined input and output interfaces, hyperparameters, and version histories as proposed in PoSyMed (*31*). *Inference-Tools* are not viewed as temporary results of individual experiments but treated as reusable research artifacts with documented provenance, access control, and publication status (*31*).
After data discovery counts are inspected and deemed sufficient by the user, a *Project* may be created based on the *DDQ*. First, a *Project* contains a set of variables that can be selected from the available $N_O$ in the *Schemas* hit by the *DDQ*. Secondly, the *Project* contains a *Workflow* in the form of a *Tool-DAG*, which allows the flow of one *Tool's* output as another *Tool's* input (*31*). Based on the *Project*, the *Researcher* can then send an *FL-Request* to all connected *Sites*. All *Sites* receive the *FL-Request* and reevaluate the *DDQ* for all cohorts. If the result is not zero after disclosure control, the *FL-Request* will be saved awaiting approval by the *Site-Owners*. After approval, the *Site* responds to the *FL-Request* with the updated count, applying the same privacy measures as with every *DDQ*. An FL *Project* run is only started once the *Project* creator starts the run. They can wait until enough data is approved for their task by the available *Sites*. Once initiated, each *Site* exports the approved patients and selected features from its local EAV database into a container-mounted dataset (Method fig. 7). Any requested features that are not available at a Site are added as null-filled columns in the case of a wide-format export, or are not added in the case of a long-format export. Any requested features that are not available at all *Sites* are dropped. The *Tools* validate the input, train locally, and exchange model weights through their FL-Net *Site* Controller via the *Platforms* Relay Server. Aggregated weights are returned for subsequent rounds, with the aggregator running at a random *Site*. Run states are synchronized with the *Platform*. Approved Tool outputs, such as the final model weights, are persisted after completion. If the approved *FL-Request* asked for the training of a public model and all participating *Sites* accepted the final model weights are uploaded to the *Platform* and can be used to package it into an *Inference-Tool*.

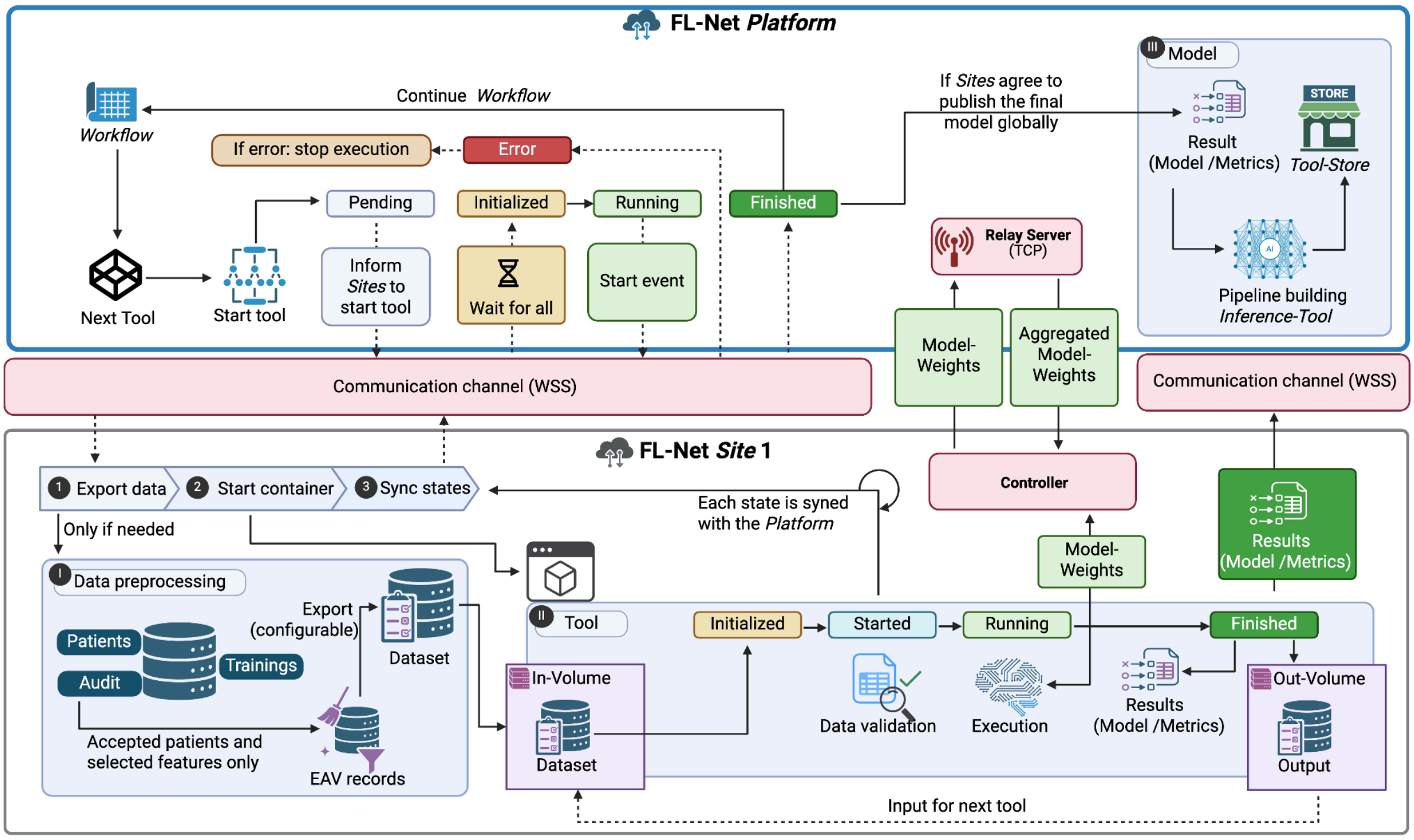

**Figure 7 - Training process in FL-Net:** Accepted patient data are prepared (1) and processed (I) fully in locally running, containerized *Tools*, while execution states are synchronized through secure websocket (WSS) (3). Once FL is started (2), only model weights are exchanged through the Controller and Relay Server channel; raw patient data remain at each *Site*. After aggregation and *Workflow* completion, model weights and metrics (III) are stored in the *Platforms* Tool-Store if participating *Site-Owners* agree to this in the *FL-Request*. These weights can be used to build an *Inference-Tool*. An error state does not reveal local logs to the *Platform*.

*FL-Tools* require federated communication. This is solved using an adapted version of the FeatureCloud communication protocol which follows a star-shaped network architecture (*16*). All *Sites* connect to the *Platform's* Relay Server which is responsible to relay all messages, doing so via custom message framing using TCP as protocol. FL algorithms encapsulated in *Tools* can send messages between *Sites* or communicate with the aggregator via the Relay Server. Similarly, the aggregator may broadcast to all *Sites* or only to specific *Sites*. To handle network instabilities, the Relay Server of FeatureCloud was extended with an in-memory message store holding messages for *Sites* not yet connected. The Controller of the *Sites* was extended with an outgoing message queue and an incoming message store that includes an automatic message ID system that automatically discards outdated messages. To build the *Inference-Tools* the *Building-Pipeline* and *Tool-Store* proposed by PoSyMed were expanded (*31*). If the final model has been approved by all the participating *Sites*, it is verified and built by the *Platform*. This *Inference-Tool* is subsequently available in the Tool-Store, can be used via the FL-Net *Platform,* or can be further customized or utilized in *Workflows* (*31*).

## Security and privacy

FL-Net addresses security and privacy by design at all relevant layers of the *Platform*. Although FL prevents the centralization of clinical data, it does not address the governance requirements of distributed research infrastructures (*48*, *49*). Therefore FL-Net is built for security and data privacy on a system level, spanning the local systems of participating *Sites*, their containerized execution environment, the network layer including FL communication, the application layer, the *DDQ* privacy, and the ML artifacts produced by *Tool* execution (Method fig. 8).

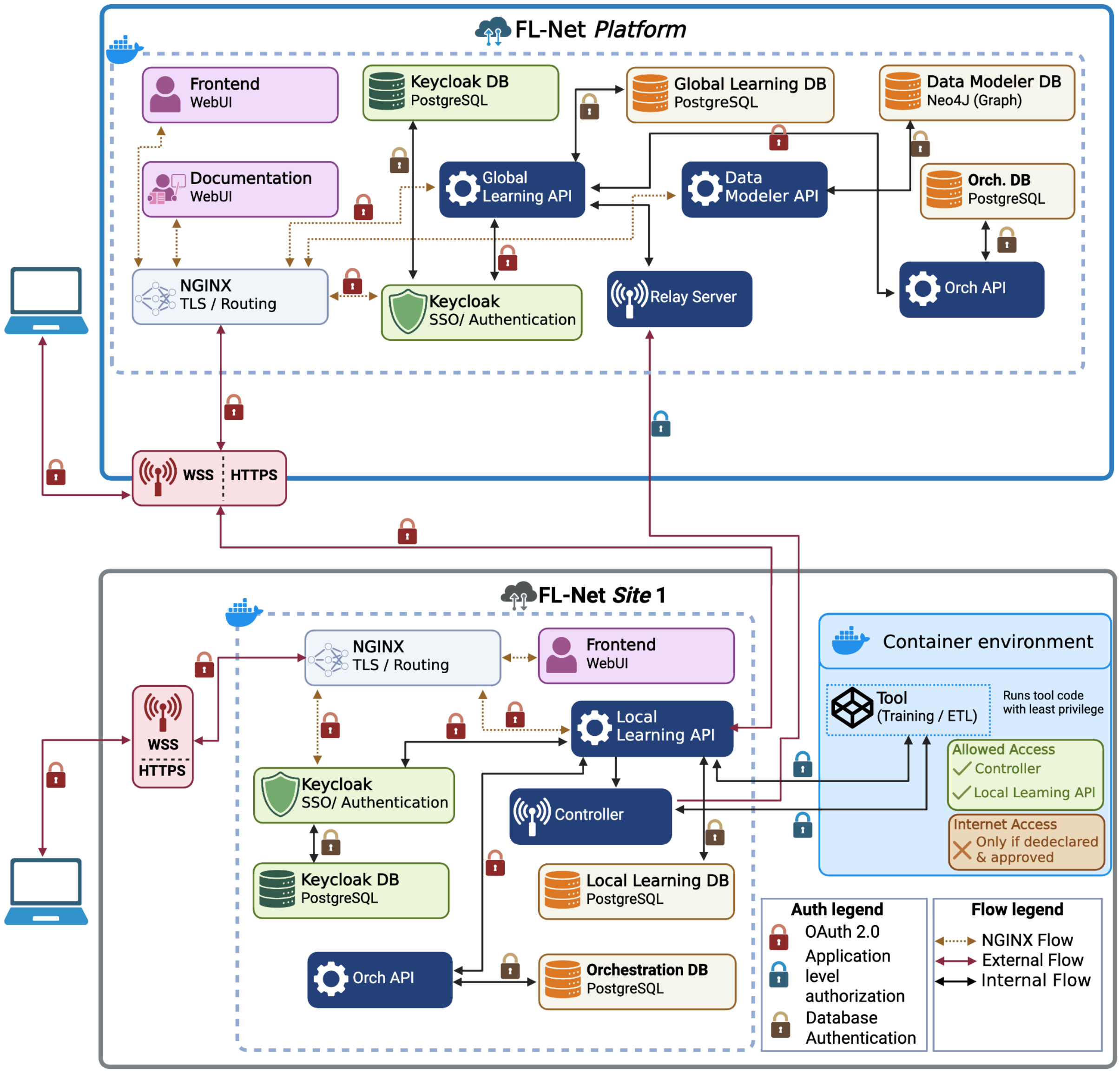

**Figure 8 - Security Architecture and Secure Communication Channels in FL-Net:** At the *Platform* level, user requests are proxied via NGINX and forwarded to the frontend, documentation, and the Global Learning API or Data Modeler API. The Orch API is not accessible from the outside.A separate Keycloak (green) instance with its own PostgreSQL database handles authentication and single sign-on. Keycloak can be replaced with any OpenID Connect server that implements a role-based system. The Global Learning API, Data Modeler API, and Orch API access only their respective assigned databases using individual credentials generated on deployment. The Global Learning API limits access on most endpoints to authenticated users only, while the Datamodeler API allows public access to ease the creation of *Schemas*. Tools may also be inspected via the Global Learning API without authentication. The Relay Server facilitates federated communication between the global infrastructure and local *Sites*, eliminating the need for direct connections between institutions. It is exposed via a separate port and uses application layer authentication and mTLS for security, with the HTTP endpoints used by the Global Learning API isolated from the Internet but not auth protected. The *Site* replicates this principle with its own NGINX, Keycloak, orchestration, and learning infrastructure, as well as institutionally controlled databases separated per service. The *Site*'s Controller receives the model weights from a *Tool*. The Controller protects the channel used by *Tools* for model weight retrieval and sending per FL *Tool* execution with application level authentication ensuring security across concurrent *Tool* executions. The Local Learning API coordinates execution of any *Tool*. All *Tools* run in a separate container environment with minimal permissions. By default, containers are allowed to access only the Local Learning API with *FL-Tools* also granted access to the Controller.

At the application level, users, services, and *Sites* are authenticated via separate Keycloak instances that use OAuth 2.0 and OpenID Connect. Role-based permissions control direct user actions. Delegated *Workflow* executions are protected via API key-based authorization.

All *Researcher* requests received via the *Platform* are checked against the permissions of the specific *Site* cohort and they may require approval by the *Site-Owners*. Automatic approvals can be set up. *DDQs* are executed locally and only return aggregated results. Minimum counts, suppression of small result sets, rounding, and *DDQ* rate limiting reduce the risk of re-identification. Alternative noise-based methods were not chosen, as they either struggle with repeated attacks or require complex algorithms to ensure the same noise is used for extremely similar queries. Thresholding and rounding offer a lower risk of disclosure with an acceptable loss of utility because DDQs are only intended to provide an approximate indication of the amount of available data. (*29*, *30*).

FL-Net provides version-specific audit reports for all *Tools* as well as provenance for *Inference-Tools*, supplementing all other protective mechanisms. Audit reports apply to specific *Tool* versions and are supplemented by usage statistics, build, and *Tool* unit test results. Each version is further automatically scanned for malware and vulnerabilities. The *Platform* links the public training artifacts to the *Tool* versions used in its specific *Project*. The *Sites* record the *Project* with exact *Tool* versions and resulting artifacts after *Project* execution. For *Inference-Tools*, release decisions, and execution metadata are also linked.

Data processing and training *Tools* run exclusively in isolated containers. By default, they can only communicate with the Local Learning API. *FL-Tools* can additionally access the Controller as this is required for FL communication. External network access is disabled by default and must be explicitly configured and approved by the *Site*. This ensures that no data can leave the *Site* without authorization. However, some *Tools* may require external access, such as extraction *Tools* populating *Sites* with data from external databases. Therefore, access to the internet is possible, although it must be enabled by the *Tool's* developer and approved by the *Site-Owners*.
FL Communication between *Sites* is running via each *Site's* Controller and the *Platforms* Relay Server, eliminating the need for direct connections between *Sites*. All connections to the *Platform* use TLS for security. The local training weights exchanged from local *Sites* to the aggregation component are additionally machine-to-machine encrypted, see S4 for further details on the encryption scheme.

## Auditing

While keeping the raw patient data inside the secure hospital infrastructure reduces the risk of transmission, all additional patient rights must be considered (*49*, *50*). The rights of data protection, withdrawal of consent, and access by data subjects, as well as the right to be forgotten, remain fully effective in federated architectures (*22*, *48*, *50*, *51*). FL-Net currently addresses this indirectly through cohort-based permissions, *DDQ* thresholds, local patient traceability, and auditable change logs To ensure traceability, every change of patient data on the local *Site*, any *DDQ* as well as any FL together with the requested patients is logged and can be audited. Any interaction with patient data can therefore be audited on a per-patient level.

# Code availability

The code for FL-Net is publicly available at the following GitHub organization: https://github.com/FedLearnNet. This organization contains the source code, documentation, and instructions for setting up and running FL-Net locally. It also contains the code for reproducing the evaluation performed for this paper at https://github.com/FedLearnNet/FL-Net-Paper. The user and technical documentation is available at https://federated-learning.net/documentation/. FL-Net is deployed at https://federated-learning.net/. To join as a *Researcher* or *Site-Owner*, please email info@mail.federated-learning.net to request account creation. To maintain platform security, all accounts are manually vetted. A future update will enable self-registration on the website, allowing you to submit your request directly online while our team conducts the approval process through the *Platform*. The FL-Net software is free to use under the Apache 2.0 License (https://www.apache.org/licenses/LICENSE-2.0.txt).

# Data availability

For the experiments testing FL-Net we used two publicly accessible clinical datasets: the Diabetes 130-US Hospitals Database, called US-130 in this article (*33*), and MIMIC-IV (*34*–*36*). The Diabetes dataset comprises 101,766 inpatient encounters collected from 130 US hospitals and integrated delivery networks between 1999 and 2008 (*33*). MIMIC-IV provides de-identified electronic health record data from patients admitted to a large tertiary-care hospital, including demographic, diagnostic, laboratory, medication and admission information (*34*–*36*). These datasets were selected because they are widely used, sufficiently heterogeneous, and accessible for reproducible benchmarking. The Diabetes 130-US Hospitals Dataset is openly available from the UCI Machine Learning Repository at: https://archive.ics.uci.edu/dataset/296/diabetes+130-us+hospitals+for+years+1999-2008.

# Funding

This project has been mainly funded through the Microb-AI-ome and dAIbetes projects. The Microb-AI-ome project has received funding from the European Union’s Horizon research and innovation programme under the Grant Agreement no. 101079777. Views and opinions expressed are, however, those of the authors only and do not necessarily reflect those of the European Union. The dAIbetes project is funded by the European Union under contract no. 101136305. The Hungarian partner is funded by the Hungarian National Research, Development, and Innovation Fund under grant number 2020-2.1.1-ED-2023-00255 (Z.B.). Views and opinions expressed are, however, those of the authors only and do not necessarily reflect those of the European Union or the Hungarian National Research, Development and Innovation Fund. Neither the European Union nor the Hungarian National Research, Development and Innovation Fund can be held responsible for them.
This project also received funding from the TEF-Health and AIM-Safe project. The TEF-Health project has received funding from the European Union's Horizon research and innovation programme under the Grant Agreement no. 101100700. Views and opinions expressed are, however, those of the authors only and do not necessarily reflect those of the European Union (B.E.). The AIM-SAFE project has received funding from the European Union's Horizon research and innovation programme under the Grant Agreement no. 101287214. Views and opinions expressed are, however, those of the authors only and do not necessarily reflect those of the European Union (B.E.).
The financial support by SBA Research (SBA-K1 NGC), a COMET Centre within the COMET – Competence Centers for Excellent Technologies Programme funded by BMIMI, BMWET, and the state of Vienna, managed by FFG, is gratefully acknowledged (R.M., A.P, and T.Š.).

# Acknowledgement

The authors only used generative AI tools for minor language editing and stylistic refinement such as checking grammar, wording, and clarity. Generative AI was not used to generate any of the scientific content, study design, analyses, results, or conclusions in this manuscript. All AI-assisted edits were critically reviewed, verified, and approved by the authors, who take full responsibility for the final content of the manuscript. All figures except fig. 5 were created using BioRender.com (*52*).

# List of Supplementary Materials

# Supplementary 1: ETL

We divided the US-130 diabetes dataset into five distinct institutional subsets and exported the data using different conventions. These conventions included different variable names, units, coding systems, languages, markers for missing values, column orders, and temporal and structural representations to create the modified US-130 dataset. Each subset was integrated via the standard FL-Net ETL chain without any code changes. A declarative connector harmonized the subsets during import into a single, common global *Schema* managed by the *Platform*. This process used a small, reusable library of cell-based functions (e.g., recording individual values, converting units/duration, normalizing missing values, and normalizing spaces and case) and row-based functions (e.g., splitting or merging columns). In addition, we used the MIMIC-IV 2.2 dataset to harmonize it with this *Schema* as well in order to demonstrate that even two completely different structures can be harmonized.
To evaluate this, we developed a common *Schema* that is used for all evaluations (fig. S1.1, RE1). The *Schema*, which was created for US-130 and MIMIC-IV, specifies which data are collected during a single inpatient stay for a patient with diabetes and comprises 50 *Schema-Nodes* ($N_S$). It should be noted that the MIMIC-IV and US-130 datasets have different *Schemas*, although their structure, *DataType-Nodes* ($N_{DT}$), and *Ontology-Nodes* ($N_O$) are identical. Due to the mapping process, some data from MIMIC-IV 2.2 was omitted, and only relevant data was mapped.

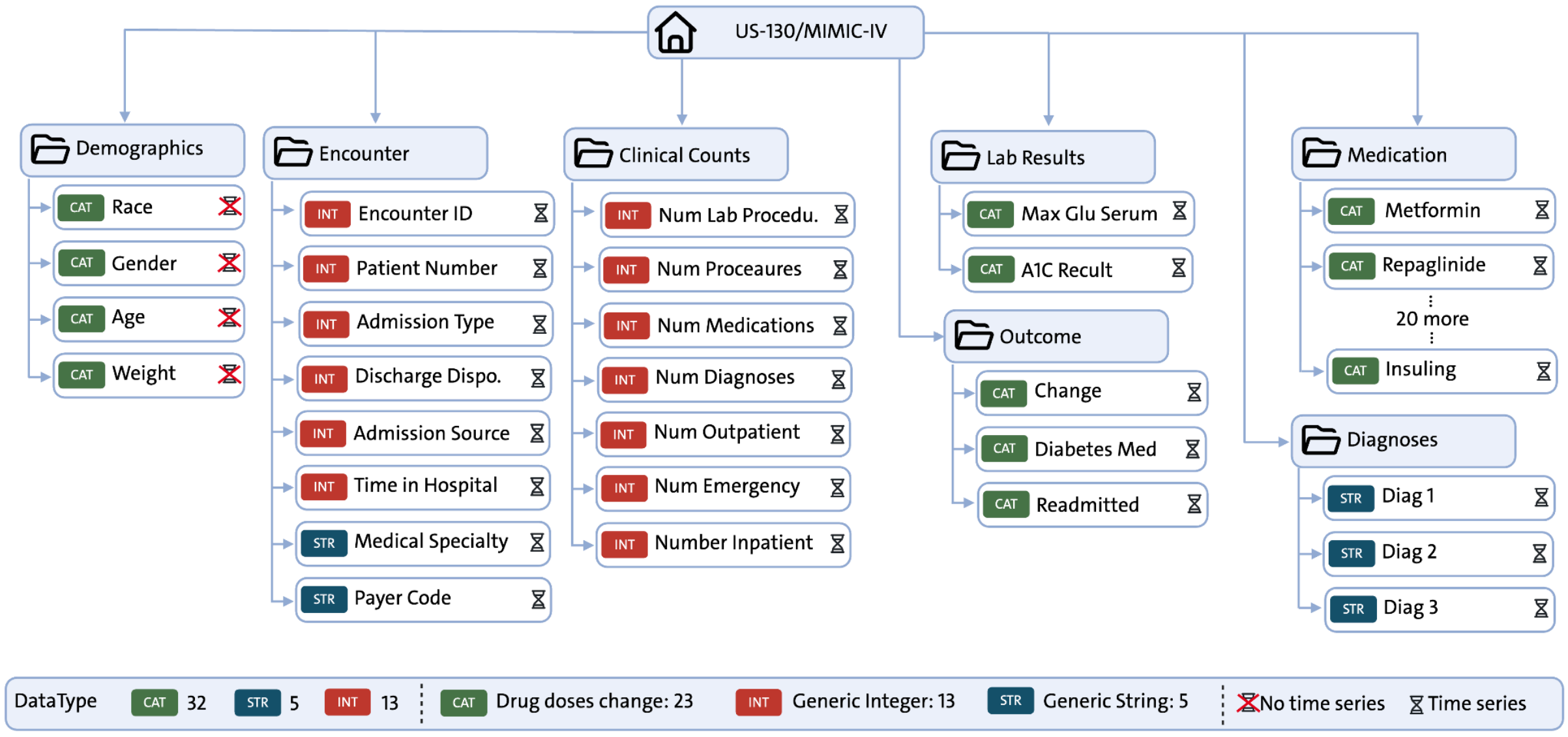

**Supplementary Figure S1.1 - Target Schema:** The *Schema* developed for US-130 and MIMIC-IV contains 50 variables that describe an inpatient stay for a patient with diabetes, grouped into seven clinical and administrative domains. Each variable is identified by the type of data it contains: CAT (categories), INT (integer), and STR (free text). The hourglass icon indicator specifies whether the variable is recorded once per patient or repeated for each hospital stay.

Four of these $N_S$ describe the patient and are recorded only once (Table S1.1). The remaining 46 are recorded at each admission, so that a patient who is admitted repeatedly is treated as a time series.

**Supplementary Table S1.1 - Variables in the evaluation scheme:** All 50 variables, grouped by the domain to which they belong. Variables with the same description are combined into a single row, and the 23 drug variables are listed in a single row, since each of them captures the same four-level dosage change for a different active ingredient formulation. Kind denotes the value type: CAT stands for a fixed set of categories, INT for an integer, and STR for free text. Values specify the permissible value range, if defined; the integer and free-text variables are not subject to such a restriction.

| Domain | Variables | Kind | Values |
|---|---|---|---|
| **Demographics** | Race | CAT | Caucasian, AfricanAmerican, Hispanic, Asian, Other |
| | Gender | CAT | Female, Male |
| | Age | CAT | 10-year bands, [0-10) … [90-100) |
| | Weight | CAT | 25-pound bands, [0-25) … >300 |
| **Encounter** | Encounter ID, Patient Number | INT | identifiers linking stays to patients |
| | Admission Type, Admission Source, Discharge Disposition | INT | integer codes; meanings not carried by the *Schema* |
| | Time in Hospital | INT | days from admission to discharge |
| | Payer Code, Medical Specialty | STR | free text |
| **Clinical Counts** | Num Lab Procedures, Num Procedures, Num Medications, Num Diagnoses | INT | volume of care during the stay |
| | Num Outpatient, Num Emergency, Num Inpatient | INT | contacts in the preceding year |
| **Diagnoses** | Diag 1, Diag 2, Diag 3 | STR | ICD-9/10 codes in fixed positional slots |
| **Lab Results** | Max Glu Serum | CAT | Norm, >200, >300, None if not measured |
| | A1C Result | CAT | Norm, >7, >8, None if not measured |

| Domain | Variables | Kind | Values |
|---|---|---|---|
| **Medication** | 16 single agents (Metformin, Insulin, Glipizide, …), 5 fixed-dose combinations, 2 fictitious | CAT | dosage change over the stay: No, Steady, Up, Down |
| **Outcome** | Change, Diabetes Med | CAT | whether diabetes medication was altered / prescribed |
| | Readmitted | CAT | <30 days, >30 days, NO |

32 variables have a fixed set of categories, 13 are integer values, and five are free-text fields. None of the variables are continuous: age, weight, and both lab results are already organized into predefined classes, so the A1C result is not a percentage but represents one of four states.
Variables that measure the same thing share a single $N_{DT}$ instead of duplicating it. The four-level dosage scale is defined once and used by all 23 medication variables, ensuring they remain comparable to one another and allowing the scale to be adjusted for all of them collectively.

# Per-Site heterogeneity and reversal evaluation

The US-130 diabetes dataset was segmented into five distinct line-by-line disjoint subsets. Subset 1 maintains the conventional UCI layout, thereby serving as the reference institution. Subsequently, the data from Subsets 2 through 5 is converted into an institutional export format by a distinct Python script for each subset (modified US-130 dataset). The corresponding connector for each institution reverses this conversion during import solely through declarative configuration (i.e., no code), so that each institution is mapped to the same common global *Schema* (Table S1.1). In order to accomplish this objective, cell- and row-based functions were utilized.
The cell-based function transforms a single column value at a time. Each cell is configured with a target column set and parameters.
For instance, the "Scale Numeric" function performs mathematical operations such as multiplication or division on a numeric value and a constant. Conversely, row functions possess the capacity to read and write an entire row, thereby enabling modifications to the column set and the generation or elimination of new rows. A case in point is the "Split Column" function, which facilitates the division of a consolidated column into multiple sections. This is the antithesis of a concatenation operation, in which fields are merged.

**Supplementary Table S1.2 - Magnitude of the simulated heterogeneity:** Each heterogeneous *Site* differs from its canonical subset in tens of thousands of individual cells, yet is reconciled by a connector with a handful of declarative steps.

| Site | Delimiter | Columns renamed | Cells changed vs canonical | Connector transformer steps |
|---|---|---|---|---|
| 2 | , | 33 | 24.929 | 11 |
| 3 | ; | 34 | 25.718 | 12 |
| 4 | , | 35 | 26.495 | 13 |
| 5 | , | 1 | 24.954 | 9 |

All *Sites* achieved a complete, *Schema-compliant* representation through configuration alone (Table S1.3). Each *Site* successfully provided the complete set of fields of the common *Schema*, including all queried variables. Value validation based on *Schema* types and allowed value domains resulted in no rejections or type conversions. Furthermore, it becomes apparent that complex cohorts such as MIMIC-IV 2.2 can be imported. The reuse of $N_{DT}$ and $N_O$ is also evaluated, since although the *Schemas* of US-130 and MIMIC-IV have different $N_S$, they share the same $P_{O\text{-}DT}$.

**Supplementary Table S1.3 - Harmonization outcomes across simulated clinical Sites**: Data from five distinct institutional subsets were integrated using the FL-Net ETL pipeline. Columns denote the *Site* identifier, the specific style of local data export, the number of columns in the raw source, the total number of records imported, the count of fields successfully mapped to the common *Schema*, and the frequency of rejected or coerced data values.

| Site | Dataset | Local export style | Source cols | Patients imported | Fields mapped | Value rejections |
|---|---|---|---|---|---|---|
| 1 | US-130 | Canonical reference | 50 | 778 | 50 | 0 |
| 2 | US-130 | Numeric coding systems (HL7 gender, race codebook, ordinal codes) | 48 | 779 | 50 | 0 |
| 3 | US-130 | Semicolon-delimited German export, length-of-stay in hours | 51 | 777 | 50 | 0 |
| 4 | US-130 | ISO-8601 duration, symbolic dosage codes, bracket-free ages | 48 | 777 | 50 | 0 |
| 5 | US-130 | Inconsistent casing/whitespace, | 51 | 780 | 50 | 0 |

| Site | Dataset | Local export style | Source cols | Patients imported | Fields mapped | Value rejections |
|---|---|---|---|---|---|---|
| | | heterogeneous missing sentinels | | | | |
| 6 | MIMIC-IV | Relational multi-table export (8 CSVs; long-format clinical events) | 73 | 299712 | 50 | 0 |

As there are no transformations at Site 1, it is not listed separately here.

## Site 2: Numeric coding-system registry

Mimics an institution that speaks in integer code systems and custom abbreviations. Columns are also renamed and reordered (Table S1.4). FL-Net maps by name, so reordering needs no configuration.

**Supplementary Table S1.4 - Reconciliation of Site 2:** This *Site* utilizes integer-based code systems and custom abbreviations. Transformation mappings demonstrate the use of the Generic Value Mapper to record standard categories, such as HL7 gender and severity grades, alongside row-level splitting of packed diagnostic codes into the shared global *Schema*.

| Variable(s) | canonical → Site export |
|---|---|
| gender→*sex* | HL7 administrative gender: *Male→1*, *Female→2* |
| race→*ethnicity_code* | custom codebook *Caucasian→R1 … Other→R5*, unknown→*R9* |
| *A1Cresult→hba1c_class* | severity grade *None→0, Norm→1, >7→2, >8→3* |
| *max_glu_serum→glu_serum_class* | grade *None→0 … >300→3* |
| *readmitted→readmission_code* | *NO→0, >30→1, <30→2* |
| *change*,*diabetesMed* | *No→0 / Ch→1*, *No→0 / Yes→1* |
| 23 medication columns (**_rx*) | ordinal dosage *No→0, Down→1, Steady→2, Up→3* |
| *weight*,*payer_code*,*medical_specialty* | missing encoded as numeric *-1* |
| *diag_1*,*diag_2*,*diag_3 → dx_bundle* | three ICD-9 codes packed as *250.83\|276\|255* |

## Site 3: European clinical information system

Mimics a non-English institution with German language, a different delimiter and unit (Table S1.5).

**Supplementary Table S1.5 - Reconciliation of Site 3:** This *Site* features a non-English institutional export using semicolon-delimited values and distinct units (e.g., hours for duration). Harmonization utilizes the Scale Numeric function for unit conversion and Generic Value Mapper for multilingual label remapping, alongside row-level operations to reconstruct age bands.

| **Variable(s)** | **canonical → Site export** |
|---|---|
| file format | comma → semicolon-delimited |
| *gender→geschlecht* | *Female→weiblich*, *Male→maennlich* |
| *race→herkunft* | German categories (*Kaukasisch*, …), unknown→*Unbekannt* |
| *time_in_hospital→verweildauer_stunden* | days → hours (×*24*) |
| *A1Cresult*,*max_glu_serum* | German labels (*Nicht gemessen*, *Ueber 8*, …) |
| *change*,*diabetesMed*,*readmitted* | German labels (*Geaendert*, *Ja*, *Binnen 30 Tagen*, …) |
| 23 medication columns (*med_**) | German dosage (*Nein*,*Reduziert*,*Stabil*,*Erhoeht*) |
| *weight*,*payer_code*,*medical_specialty* | missing encoded as *k.A.* |
| *age → alter_von*,*alter_bis* | age band *[50-60)* split on - into *[50* , *60)* |

## Site 4: Symbolic / temporal-encoding EHR

Mimics an institution that uses typographic symbols, terse tokens and an ISO-8601 temporal encoding. Columns are renamed and reordered (Table S1.6).

**Supplementary Table S1.6 - Reconciliation of Site 4** This *Site* employs typographic symbols and ISO-8601 temporal encoding. The connector utilizes the Duration To Days function for temporal conversion and specific row-level operations to fuse encounter codes, alongside Generic Value Mapping for symbol-to-canonical value reconciliation.

| **Variable(s)** | **canonical → Site export** |
|---|---|
| *time_in_hospital→StayDuration* | integer days → ISO-8601 duration (*3→P3D*) |
| 23 medication columns (*Rx_**) | dosage symbols *No→-, Steady→=, Up→+, Down→x* |
| *age→AgeRange* | bracket-free ranges *[0-10)→0-9*, … |
| *gender→Sex* | *Female→F*, *Male→M* |

| Variable(s) | canonical → Site export |
|---|---|
| *race→RaceCode* | upper codes *CAUC/AFAM/HISP/ASIA/OTHR*, unknown→*UNK* |
| *A1Cresult*,*max_glu_serum* | terse tokens *NA/WNL/H1/H2*, *NA/WNL/G2/G3* |
| *change*,*diabetesMed*,*readmitted* | *0/1*, *false/true*, *N/L/E* |
| *weight*,*payer_code*,*medical_specialty* | missing encoded as *.* |
| *admission_type_id*,*discharge_disposition_id*,*admission_source_id* → *EncounterCodes* | three codes fused as *1-1-7* |

## Site 5: Unstructured text extract

Mimics a spreadsheet dump: no code systems, just dirtiness, inconsistent casing, stray whitespace, and missing-value placeholders (Table S1.6). This stresses cleaning built-ins.
The dataset exhibits characteristics reminiscent of a spreadsheet dump, characterized by the absence of code systems, the presence of dirtiness, inconsistent casing, stray whitespace, and a proliferation of missing-value sentinels (Table S1.7). This underscores the importance of meticulous cleaning of medical data.

**Supplementary Table S1.7: Reconciliation of Site 5.** This *Site* mimics unformatted spreadsheet output with inconsistent casing, whitespace variability, and heterogeneous missing-value sentinels. The connector primarily exercises data cleaning built-ins, including Trim And Case, Regex Replace, and Normalize Missing, to align the dataset with the canonical structure.

| Variable(s) | canonical → Site export |
|---|---|
| *gender*, *A1Cresult*, *max_glu_serum*, *change*, *diabetesMed*, 23 meds | lower-cased and whitespace-padded (*Steady*→" *steady* ") |
| *readmitted* | comparators spaced out (*<30* → " *< 30* ") |
| *race* | whitespace-padded; missing as *NULL*/*N/A* |
| *weight*,*payer_code*,*medical_specialty* | distinct sentinels *-99*, *NULL*, *unknown* |
| *age* → *age_low*,*age_high* | age band split on - |

## Site 6: MIMIC-IV

The connector for the MIMIC-IV 2.2 dataset executes the following 51 transformers in numerical order (Table S1.8). Patient-level functions place encounter-derived values on the admission row identified by *hadm_id*; cell functions then record or bin those values in place.

Steps 26-48 are separate configured transformers, one for each medication $N_S$, even though they use the same reusable dosage-trend function.

**Supplementary Table S1.8: Complete connector transformation sequence.** Every configured transformer is listed separately with its operation and target.

| Step | Transformer and target | Explanation |
|---|---|---|
| 1 | Patient Value Aggregate - Patient Number | Copies the first subject_id for each admissions::hadm_id to patient_nbr and writes it on the matching admission row, giving every visit the *Schema's* patient-number field. |
| 2 | Generic Value Mapper - Gender | Maps patients.gender in place from M/F to the *Schema* categories Male/Female. |
| 3 | Generic Value Mapper - Race | Collapses 33 MIMIC race labels into the five *Schema* categories Caucasian, AfricanAmerican, Hispanic, Asian, and Other; the reduction is lossy but makes the cohorts comparable. |
| 4 | Numeric Range Group Mapper - Age | Bins continuous anchor_age into the *Schema's* ten-year age bands from [0-10) through [90-100). |
| 5 | Patient Value Aggregate - Weight | Finds OMR rows whose result_name is Weight (Lbs), selects the latest result_value by chartdate for each patient, and writes one weight_lbs value to the patient anchor row. |
| 6 | Numeric Range Group Mapper - Weight | Bins weight_lbs into 25-pound target bands from [0-25) through [275-300), with values above the configured maximum represented as >300. |
| 7 | Generic Value Mapper - Admission Type | Maps MIMIC admission_type labels to the target integer admission-type codes, including emergency, urgent, elective, and observation classes. |
| 8 | Generic Value Mapper - Admission Source | Maps admission_location labels to target admission-source codes, including referrals, transfers, emergency-room, self-referral, and *procedure-Site* sources. |
| 9 | Generic Value Mapper - Discharge Disposition | Maps discharge_location labels to target discharge-disposition codes, including home, facility, home-health, death, hospice, rehabilitation, and long-term care. |

| Step | Transformer and target | Explanation |
|---|---|---|
| 10 | Date Difference - Time in Hospital | Subtracts admittime from dischtime, rounds half-up to whole days, and writes time_in_hospital for the admission. |
| 11 | Patient Value Aggregate - Num Lab Procedures | Counts labevents rows within each labevents::hadm_id and writes num_lab_procedures to the corresponding admission. This runs before step 21 removes irrelevant lab rows. |
| 12 | Patient Value Aggregate - Num Procedures | Counts procedures_icd::icd_code rows per hadm_id and writes num_procedures to the matching admission. |
| 13 | Patient Value Aggregate - Num Medications | Counts distinct prescriptions.drug names per hadm_id and writes num_medications to the matching admission. |
| 14 | Patient Value Aggregate - Number Diagnoses | Counts diagnoses_icd::icd_code rows per hadm_id and writes number_diagnoses to the matching admission. |
| 15 | Patient Prior Event Count - Number Outpatient | Counts earlier admissions in the preceding 365 days whose admission_type is AMBULATORY OBSERVATION, DIRECT OBSERVATION, or OBSERVATION ADMIT; writes the result as number_outpatient. |
| 16 | Patient Prior Event Count - Number Emergency | Counts earlier admissions in the preceding 365 days whose admission_type is EW EMER., DIRECT EMER., EU OBSERVATION, or URGENT; writes number_emergency. |
| 17 | Patient Prior Event Count - Number Inpatient | Counts all earlier admissions for the patient within the preceding 365 days and writes number_inpatient. |
| 18 | Patient Value Aggregate - Diag 1 | Selects the first diagnosis code with diagnoses_icd::seq_num = 1 in each hadm_id and writes it to diag_1. |
| 19 | Patient Value Aggregate - Diag 2 | Selects the first diagnosis code with diagnoses_icd::seq_num = 2 in each hadm_id and writes it to diag_2. |

| Step | Transformer and target | Explanation |
|---|---|---|
| 20 | Patient Value Aggregate - Diag 3 | Selects the first diagnosis code with diagnoses_icd::seq_num = 3 in each hadm_id and writes it to diag_3. |
| 21 | Filter Rows By Value - row set | Keeps labevents rows only when itemid is 50852, 50931, or 50809, while retaining rows from other tables that have no labevents::itemid. This reduces the event stream before downstream steps. |
| 22 | Patient Value Aggregate - A1C Result | For each admission, takes the maximum labevents::valuenum for itemid 50852 (hemoglobin A1c) and writes the numeric intermediate a1c_result. |
| 23 | Numeric Threshold Class - A1C Result | Converts a1c_result to target categories: values below 7 become Norm, values from 7 to below 8 become >7, values at least 8 become >8, and no measurement becomes None. |
| 24 | Patient Value Aggregate - Max Glu Serum | For each admission, takes the maximum labevents::valuenum across itemids 50931 and 50809 (glucose) and writes the numeric intermediate max_glu_serum. |
| 25 | Numeric Threshold Class - Max Glu Serum | Converts max_glu_serum to target categories: values below 200 become Norm, values from 200 to below 300 become >200, values at least 300 become >300, and no measurement becomes None. |
| 26 | Patient Dosage Trend - Metformin | Matches metformin, glucophage, fortamet, riomet, glumetza, compares first and last dose, writes Metformin as No, Steady, Up, or Down. |
| 27 | Patient Dosage Trend - Repaglinide | Matches repaglinide, prandin, compares first and last dose, writes Repaglinide as No, Steady, Up, or Down. |
| 28 | Patient Dosage Trend - Nateglinide | Matches nateglinide, starlix, compares first and last dose, writes Nateglinide as No, Steady, Up, or Down. |
| 29 | Patient Dosage Trend - Chlorpropamide | Matches chlorpropamide, diabinese, compares first and last dose, writes Chlorpropamide as No, Steady, Up, or Down. |

| Step | Transformer and target | Explanation |
|---|---|---|
| 30 | Patient Dosage Trend - Glimepiride | Matches glimepiride, amaryl, compares first and last dose, writes Glimepiride as No, Steady, Up, or Down. |
| 31 | Patient Dosage Trend - Acetohexamide | Matches acetohexamide, dymelor, compares first and last dose, writes Acetohexamide as No, Steady, Up, or Down. |
| 32 | Patient Dosage Trend - Glipizide | Matches glipizide, glucotrol, compares first and last dose, writes Glipizide as No, Steady, Up, or Down. |
| 33 | Patient Dosage Trend - Glyburide | Matches glyburide, glibenclamide, diabeta, micronase, glynase, compares first and last dose, writes Glyburide as No, Steady, Up, or Down. |
| 34 | Patient Dosage Trend - Tolbutamide | Matches tolbutamide, orinase, compares first and last dose, writes Tolbutamide as No, Steady, Up, or Down. |
| 35 | Patient Dosage Trend - Pioglitazone | Matches pioglitazone, actos, compares first and last dose, writes Pioglitazone as No, Steady, Up, or Down. |
| 36 | Patient Dosage Trend - Rosiglitazone | Matches rosiglitazone, avandia, compares first and last dose, writes Rosiglitazone as No, Steady, Up, or Down. |
| 37 | Patient Dosage Trend - Acarbose | Matches acarbose, precose, compares first and last dose, writes Acarbose as No, Steady, Up, or Down. |
| 38 | Patient Dosage Trend - Miglitol | Matches miglitol, glyset, compares first and last dose, writes Miglitol as No, Steady, Up, or Down. |
| 39 | Patient Dosage Trend - Troglitazone | Matches troglitazone, rezulin, compares first and last dose, writes Troglitazone as No, Steady, Up, or Down. |
| 40 | Patient Dosage Trend - Tolazamide | Matches tolazamide, tolinase, compares first and last dose, writes Tolazamide as No, Steady, Up, or Down. |
| 41 | Patient Dosage Trend - Examide | Matches examide, compares first and last dose, writes Examide as No, Steady, Up, or Down. |

| Step | Transformer and target | Explanation |
|---|---|---|
| 42 | Patient Dosage Trend - Citoglipton | Matches citoglipton, compares first and last dose, writes Citoglipton as No, Steady, Up, or Down. |
| 43 | Patient Dosage Trend - Insulin | Matches insulin, humalog, novolog, lantus, levemir, humulin, novolin, glargine, lispro, aspart, detemir, degludec, glulisine, toujeo, tresiba, compares first and last dose, writes Insulin as No, Steady, Up, or Down. |
| 44 | Patient Dosage Trend - Glyburide-Metformin | Matches glucovance, glyburide-metformin, compares first and last dose, writes Glyburide-Metformin as No, Steady, Up, or Down. |
| 45 | Patient Dosage Trend - Glipizide-Metformin | Matches metaglip, glipizide-metformin, compares first and last dose, writes Glipizide-Metformin as No, Steady, Up, or Down. |
| 46 | Patient Dosage Trend - Glimepiride-Pioglitazone | Matches duetact, glimepiride-pioglitazone, compares first and last dose, writes Glimepiride-Pioglitazone as No, Steady, Up, or Down. |
| 47 | Patient Dosage Trend - Metformin-Rosiglitazone | Matches avandamet, metformin-rosiglitazone, compares first and last dose, writes Metformin-Rosiglitazone as No, Steady, Up, or Down. |
| 48 | Patient Dosage Trend - Metformin-Pioglitazone | Matches actoplus, metformin-pioglitazone, compares first and last dose, writes Metformin-Pioglitazone as No, Steady, Up, or Down. |
| 49 | Column Set Indicator - Change | Checks all 23 medication outputs; if any is Up or Down, writes med_change = Ch, otherwise writes No. |
| 50 | Column Set Indicator - Diabetes Med | Checks all 23 medication outputs; if any is not No, writes diabetes_med = Yes, otherwise writes No. |
| 51 | Readmission Interval Class - Readmitted | Orders admissions within each patient and classifies the interval after discharge as readmitted <30, >30, or NO when there is no later admission. |

# Supplementary 2: Data Discovery Query and disclosure control

A *DDQ* can be composed by a *Researcher* via the frontend. A *DDQ* is a list of *DDQ* items, where each item represents a condition (fig. S2.1A).

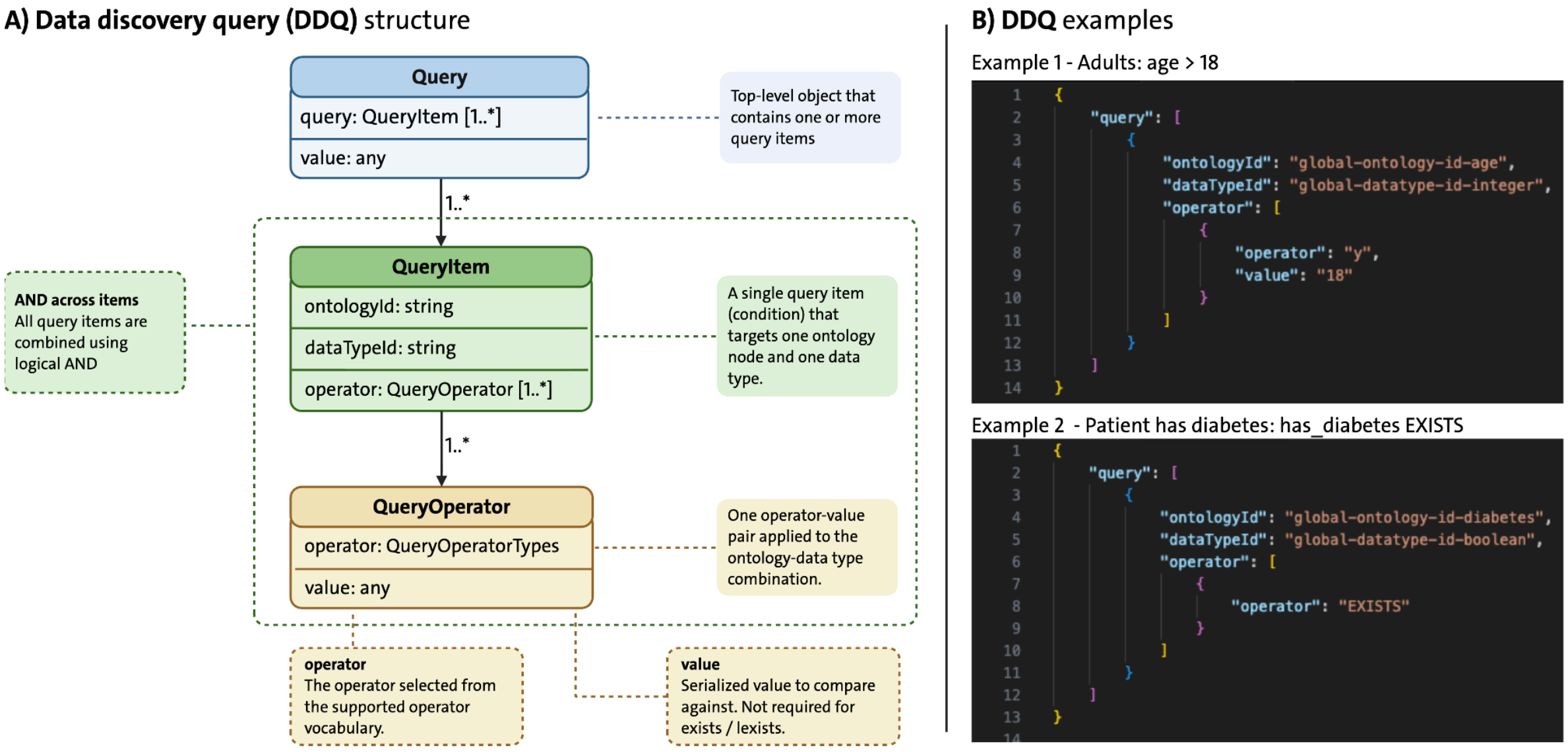

**Supplementary Figure S2.1 - DDQ Structure:** A) FL-Net expresses *DDQs* independently of *Schemas* through available combinations of $N_O$ and $N_{DT}$ identifiers. A *DDQ* consists of one or more *DDQ* items that are combined by logical AND. Each *DDQ* item targets a specific $P_{O\text{-}DT}$ and contains one or more operator–value conditions defined by the supported *DDQ* vocabulary. B) Example serialized *DDQ* illustrates the *Schema* for a numerical constraint (age > 18) and a concept-presence constraint (patient has diabetes). Because a *DDQ* addresses shared semantic $N_O$ and $N_{DT}$ definitions rather than cohort-specific $N_S$, the same *DDQ* can be evaluated across heterogeneous *Schemas* that resolve to the corresponding $P_{O\text{-}DT}$.

Each *DDQ* item contains the globally available *OntologyId* (the id of the specific $N_O$) and *DataTypeId* (the id of the specific $N_{DT}$). Furthermore, each item contains an operator and a value. The operators are <, <=, >, >=, =, !=, EXISTS, NOT EXISTS, CONTAINS, STARTS WITH, ENDS WITH and REGEX. The system limits the operators depending on the $N_{DT}$, e.g. REGEX is not available for a numerical $N_{DT}$. The values are also limited based on the $N_{DT}$, e.g. for a categorical $N_{DT}$ the value selection is a drop-down of the allowed categories. Local *Sites* resolve the global references to their respective *Schema* structure and compile stand-alone patient id count SQL queries against the local EAV data store for any cohort that contains all $P_{O\text{-}DT}$ from the *DDQ*. A simple example *DDQ* would be the $N_O$ age, the $N_{DT}$ integer between 0 and 120, the operator >, and the value 18 to only include adults in data discovery (fig. S2.1B).
In this process, cohort-specific permissions are checked before running any local *DDQ* as well as before results are returned. These permissions may be specific to a *Platform* user based on the user id sent by the *Platform* with the *DDQ*, or they can be set for any *Platform* user. Cohorts are only queried if the *Platform* user is permitted and if the cohort-specific *DDQ* cooldown since the last answered *DDQ* has passed. Resulting counts are further modified. Where minimum patient count thresholds per cohort are enforced locally, the count results of multiple cohorts are summed up and the reported counts are additionally rounded. Any *DDQ* and affected patients are logged locally for auditing.

## S2.1 Queries and local execution (RE4)

A *DDQ* is issued from the global *Platform* as an ordered list of *DDQ* items. Each item targets one ontology concept and one or more DataType. The ontology identifier serves as the concept's global reference (e.g. C0034510 (Race) or C0600032 (Readmitted)). Critically, the *DDQ* contains no SQL and no *Site-specific Schema* information, the *DDQ* is sent to all *Sites* and the instruction set is identical for all *Sites*.

Each participating *Site* compiles the *DDQ* locally and deterministically against its harmonized data store. This store is structured as a long-format, EAV table.

For each *DDQ* item, the *Site* executes a three-step resolution:

1. **Map:** Resolve the OntologyId to local $N_S$ via the local mapping. Detect the $N_S$ with the requested $N_{DT}$ and reject the *DDQ* if the requested combination is not available.
2. **Cast:** Identify the appropriate value column based on the $N_{DT}$ (e.g., CATEGORICAL or STRING map to value_string; INT maps to value_int).
3. **Emit:** Generate a" SELECT DISTINCT patient_id" statement for each resolved $P_{O\text{-}DT}$.

$P_{O\text{-}DT}$ corresponding to the same $N_O$ are combined using UNION operations, while distinct items in the *DDQ* list are combined using INTERSECT. This ensures that the final result counts only the distinct patients satisfying all criteria.

**Example.** A *DDQ* list targeting the ontology C0034510 (Race) equals the value "Caucasian" (C0034510) of the *Datatype* Race (Snomed) AND the ontology "Female" (C4019318) equals the value true of the *Datatype* generic bool compiles locally to:
SELECT DISTINCT patient_id FROM (SELECT DISTINCT patient_id FROM patient_data WHERE schema_node_id = <Race_Schemanode_id> and value_string = 'Caucasian') INTERSECT (SELECT DISTINCT patient_id FROM patient_data WHERE schema_node_id = <Gender_Female_id> and value_bool = true).

Similarly, a numeric *DDQ* item requesting the ontology "length of stay" (C0420382) with the operator > 7 on the $N_{DT}$ days compiles to a filter where *SchemaNodeId* = <los_node> and value_int > 7. Because only *Site-specific* variables are resolved deterministically from the global ontology identifier, identical *DDQ* inputs reliably yield identical logical cohorts across all *Sites*, fulfilling the semantic *DDQ* requirement (RE4).

## S2.2 Disclosure-control policy (RE2)

Before any count is returned, each *Site* applies, in order:

1. **Access control:** The requesting identity must hold a cohort-scoped permission with *DDQ* access; non-permitted cohorts are not queried.
2. **Timing control:** Each *DDQ* specifies how much time must pass between queries (*DDQ* retry time). If a *DDQ* was executed on the cohort and another *DDQ* is requested before this time has passed, the cohort is excluded from this subsequent *DDQ*.

3. **Per-cohort minimum-sample threshold:** A cohort contributes only if the number of matching patients is at least the cohort's configured sample threshold *k*.
4. **Rounding:** The total count *n* over all queried cohorts is rounded up to one order of magnitude below its own.
5. **Traceability:** The matching patient ids and further meta-information, such as the querying user, the time of the *DDQ*, and the *DDQ* itself, are recorded locally (never transmitted). This ensures that an approved downstream run can be authorized against exactly this cohort. Furthermore, this is required for auditing purposes.

The released, rounded count is the only information that leaves the clinic.
Rounding/suppression examples (default k = 100):

| **raw count n** | **released** | **note** |
|---|---|---|
| 47 | 0 | < k (e.g. 100) → suppressed |
| 123 | 130 | round up to nearest 10 |
| 1,234 | 1,300 | round up to nearest 100 |
| 20,148 | 21,000 | round up to nearest 1,000 |
| 640,148 | 641,000 | round up to nearest 10,000 |

This is the standard statistical disclosure control pattern (small count suppression and rounded released counts). The thresholds are *Site* specific settings and are reported here so that the privacy-utility behavior is fully specified and reproducible. We reuse the identical rounding when reporting discovered patient sizes as metrics, ensuring that no exact count ever leaves a *Site*.

## S2.3 Evaluation queries

We evaluated *DDQ* and disclosure control with 13 predefined *DDQs*. Each *DDQ* was sent from the global FL-Net *Platform* to all 130 simulated clinics of the UCI Diabetes US-130 Hospitals dataset (S1). Queries were expressed through ontology-based predicates, executed locally on distinct patient identifiers, and released only after the local disclosure-control policy had been applied. All *Sites* set the same *DDQ* retry time and queries were executed so that the *DDQ* retry time passed.

**all_female**
Question: How many distinct patients are female?
Predicate: Gender = Female.

**race_caucasian**
Question: How many distinct patients are Caucasian?
Predicate: Race = Caucasian.

**a1c_not_measured**
Question: How many distinct patients have no measured A1C result?
Predicate: A1C result = None.

**on_diabetes_meds**
Question: How many distinct patients received diabetes medication?
Predicate: DiabetesMed = Yes.

**metformin_steady**
Question: How many distinct patients were on steady metformin?
Predicate: Metformin = Steady.

**insulin_steady**
Question: How many distinct patients were on steady insulin?
Predicate: Insulin = Steady.

**long_stay**
Question: How many distinct patients stayed in hospital for more than seven days?
Predicate: Time in Hospital > 7.

**readmit_30**
Question: How many distinct patients were readmitted within 30 days?
Predicate: Readmitted = <30.

**caucasian_female**
Question: How many distinct patients are both Caucasian and female?
Predicate: Race = Caucasian AND Gender = Female.

**a1c_high_metformin_up**
Question: How many distinct patients had A1C >8 and an increased metformin dose?
Predicate: A1C result = >8 AND Metformin = Up.

**race_asian**
Question: How many distinct patients are Asian?
Predicate: Race = Asian.

**aa_insulin_up**
Question: How many distinct African American patients had an increased insulin dose?
Predicate: Race = AfricanAmerican AND Insulin = Up.

**asian_male_a1c8**
Question: How many distinct Asian male patients had A1C >8?
Predicate: Race = Asian AND Gender = Male AND A1C result = >8.

## S2.4 Query Results

Query reproducibility and disclosure control were evaluated across all *Sites* and predefined *DDQs* (Tables S2.1–S2.2). Repeated execution produced identical released counts across all three runs, demonstrating deterministic query behavior. At the same time, disclosure controls modified the returned results where required: discovered patient counts were rounded before release, whereas cohorts below the configured disclosure threshold were suppressed entirely. Together, these results show that FL-Net provides reproducible *DDQs* while consistently enforcing privacy-preserving release rules at the local *Sites*.

**Supplementary Table S2.1 - Reproducibility:** For each *Site* × logical *DDQ*: the released count of repeat 1, 2, and 3, and whether all three are identical (expected: yes for every row). For Sites 1–3, the US-130 dataset was used, while *Site* 4 used MIMIC-IV 2.2. All data were imported as described in S1. The global total would therefore be the sum of all four *Sites* combined.

| Site | DDQ Key | Released r1 | Released r2 | Released r3 | Identical |
|---|---|---|---|---|---|
| 1 | all_female | 410 | 410 | 410 | Yes |
| 1 | race_caucasian | 580 | 580 | 580 | Yes |
| 1 | a1c_not_measured | 640 | 640 | 640 | Yes |
| 1 | on_diabetes_meds | 580 | 580 | 580 | Yes |
| 1 | metformin_steady | 150 | 150 | 150 | Yes |
| 1 | insulin_steady | 220 | 220 | 220 | Yes |
| 1 | long_stay | 130 | 130 | 130 | Yes |
| 1 | readmit_30 | 0 | 0 | 0 | Yes |
| 1 | caucasian_female | 290 | 290 | 290 | Yes |
| 1 | a1c_high_metformin_up | 0 | 0 | 0 | Yes |
| 1 | race_asian | 0 | 0 | 0 | Yes |
| 1 | aa_insulin_up | 0 | 0 | 0 | Yes |
| 1 | asian_male_a1c8 | 0 | 0 | 0 | Yes |
| 2 | all_female | 420 | 420 | 420 | Yes |
| 2 | race_caucasian | 590 | 590 | 590 | Yes |
| 2 | a1c_not_measured | 640 | 640 | 640 | Yes |
| 2 | on_diabetes_meds | 600 | 600 | 600 | Yes |
| 2 | metformin_steady | 150 | 150 | 150 | Yes |
| 2 | insulin_steady | 250 | 250 | 250 | Yes |
| 2 | long_stay | 130 | 130 | 130 | Yes |
| 2 | readmit_30 | 0 | 0 | 0 | Yes |
| 2 | caucasian_female | 310 | 310 | 310 | Yes |
| 2 | a1c_high_metformin_up | 0 | 0 | 0 | Yes |
| 2 | race_asian | 0 | 0 | 0 | Yes |
| 2 | aa_insulin_up | 0 | 0 | 0 | Yes |
| 2 | asian_male_a1c8 | 0 | 0 | 0 | Yes |
| 3 | all_female | 390 | 390 | 390 | Yes |
| 3 | race_caucasian | 570 | 570 | 570 | Yes |
| 3 | a1c_not_measured | 640 | 640 | 640 | Yes |
| 3 | on_diabetes_meds | 590 | 590 | 590 | Yes |
| 3 | metformin_steady | 140 | 140 | 140 | Yes |
| 3 | insulin_steady | 250 | 250 | 250 | Yes |
| 3 | long_stay | 120 | 120 | 120 | Yes |
| 3 | readmit_30 | 0 | 0 | 0 | Yes |
| 3 | caucasian_female | 280 | 280 | 280 | Yes |
| 3 | a1c_high_metformin_up | 0 | 0 | 0 | Yes |
| 3 | race_asian | 0 | 0 | 0 | Yes |
| 3 | aa_insulin_up | 0 | 0 | 0 | Yes |
| 3 | asian_male_a1c8 | 0 | 0 | 0 | Yes |
| 4 (MIMIC-IV) | all_female | 160000 | 160000 | 160000 | Yes |
| 4 (MIMIC-IV) | race_caucasian | 130000 | 130000 | 130000 | Yes |
| 4 (MIMIC-IV) | a1c_not_measured | 180000 | 180000 | 180000 | Yes |
| 4 (MIMIC-IV) | on_diabetes_meds | 57000 | 57000 | 57000 | Yes |
| 4 (MIMIC-IV) | metformin_steady | 7800 | 7800 | 7800 | Yes |
| 4 (MIMIC-IV) | insulin_steady | 42000 | 42000 | 42000 | Yes |
| 4 (MIMIC-IV) | long_stay | 44000 | 44000 | 44000 | Yes |
| 4 (MIMIC-IV) | readmit_30 | 41000 | 41000 | 41000 | Yes |
| 4 (MIMIC-IV) | caucasian_female | 64000 | 64000 | 64000 | Yes |
| 4 (MIMIC-IV) | a1c_high_metformin_up | 250 | 250 | 250 | Yes |
| 4 (MIMIC-IV) | race_asian | 7700 | 7700 | 7700 | Yes |
| 4 (MIMIC-IV) | aa_insulin_up | 2500 | 2500 | 2500 | Yes |
| 4 (MIMIC-IV) | asian_male_a1c8 | 0 | 0 | 0 | Yes |

**Supplementary Table S2.2 Disclosure control:.** For each *Site* × logical *DDQ*: raw vs. released count, the rounding delta, and whether the discovered patient count was suppressed.

| Site | DDQ Key | Raw | Released | Δ (Released − Raw) | Suppressed |
|---|---|---|---|---|---|
| 1 | all_female | 401 | 410 | 9 | No |
| 1 | race_caucasian | 577 | 580 | 3 | No |
| 1 | a1c_not_measured | 637 | 640 | 3 | No |
| 1 | on_diabetes_meds | 580 | 580 | 0 | No |
| 1 | metformin_steady | 150 | 150 | 0 | No |
| 1 | insulin_steady | 218 | 220 | 2 | No |
| 1 | long_stay | 124 | 130 | 6 | No |
| 1 | readmit_30 | 87 | 0 | -87 | Yes |
| 1 | caucasian_female | 288 | 290 | 2 | No |
| 1 | a1c_high_metformin_up | 2 | 0 | -2 | Yes |
| 1 | race_asian | 2 | 0 | -2 | Yes |
| 1 | aa_insulin_up | 10 | 0 | -10 | Yes |
| 1 | asian_male_a1c8 | 0 | 0 | 0 | No |
| 2 | all_female | 415 | 420 | 5 | No |
| 2 | race_caucasian | 582 | 590 | 8 | No |
| 2 | a1c_not_measured | 636 | 640 | 4 | No |
| 2 | on_diabetes_meds | 600 | 600 | 0 | No |
| 2 | metformin_steady | 149 | 150 | 1 | No |
| 2 | insulin_steady | 249 | 250 | 1 | No |
| 2 | long_stay | 130 | 130 | 0 | No |
| 2 | readmit_30 | 95 | 0 | -95 | Yes |
| 2 | caucasian_female | 304 | 310 | 6 | No |
| 2 | a1c_high_metformin_up | 2 | 0 | -2 | Yes |
| 2 | race_asian | 4 | 0 | -4 | Yes |
| 2 | aa_insulin_up | 18 | 0 | -18 | Yes |
| 2 | asian_male_a1c8 | 0 | 0 | 0 | No |
| 3 | all_female | 389 | 390 | 1 | No |
| 3 | race_caucasian | 561 | 570 | 9 | No |
| 3 | a1c_not_measured | 640 | 640 | 0 | No |
| 3 | on_diabetes_meds | 590 | 590 | 0 | No |
| 3 | metformin_steady | 138 | 140 | 2 | No |
| 3 | insulin_steady | 242 | 250 | 8 | No |
| 3 | long_stay | 112 | 120 | 8 | No |
| 3 | readmit_30 | 86 | 0 | -86 | Yes |
| 3 | caucasian_female | 278 | 280 | 2 | No |
| 3 | a1c_high_metformin_up | 1 | 0 | -1 | Yes |
| 3 | race_asian | 7 | 0 | -7 | Yes |
| 3 | aa_insulin_up | 15 | 0 | -15 | Yes |
| 3 | asian_male_a1c8 | 0 | 0 | 0 | No |

Reporting raw and released side-by-side renders the privacy–utility trade-off of the *DDQ* layer transparent: every released count is ≥ its raw value (conservative upward rounding) or 0 (suppression), and no released count discloses an exact patient number.

# Supplementary 3: Federated neural-network training for 30-day hospital readmission

The object of this experiment was to evaluate whether FL with FL-Net could improve readmission prediction within 30 days compared to institutional training while equaling the performance of a centralized, pooled reference model. This shall generally prove that the FL framework used within FL-Net is correctly implemented.

All experiments used the Diabetes US-130 Hospitals Dataset (*1*). This dataset comprises 101,766 inpatient encounters recorded at 130 US hospitals and integrated delivery networks between 1999 and 2008 (*1*). Their institutional and temporal structure also makes them suitable for simulating distributed clinical environments under controlled conditions. Importantly, the "130" in the Diabetes dataset refers to the hospitals contributing to the original data collection rather than the number of federation *Sites* used in our experiments, which varied independently across 3, 5, 10, 25, 33, and 50 *Sites*. The prediction target is hospital readmission within 30 days of discharge. This is a binary classification task in which the three-level readmission field in the dataset is mapped to a positive class for readmission in less than 30 days and a negative class for readmission after 30 days (*1*). We chose this dataset for several reasons: it is public, widely used, and naturally compatible with FL as it is originally gathered from distributed sources, which allows others to reproduce the full pipeline without restricted-access data.

First, isolated training at a single *Site* was used as the baseline. In this condition, each participating *Site* trained a model using only its own data and was evaluated using a validation split held out from the same *Site*. This corresponds to the status quo, in which institutions cannot benefit from the observations of other centers. Second, federated FL-Net training was used as a privacy-compliant, collaborative condition. In this condition, each institution trained locally and exchanged only model parameters with the federation backend. The institutions also received an aggregated global model for the next round. Third, a centralized, pooled training was used as the upper reference. In this condition, all training datasets were combined into a single, centralized dataset. A single model was then trained and evaluated using the held-out, unknown evaluation dataset. Therefore, the federated condition was evaluated based on its improvement over the isolated lower reference and its proximity to the pooled upper reference.

To evaluate FL-Net under controlled yet realistic conditions, we emulated a multi-institution federation on a single server with a Docker-in-Docker deployment (*2*). Each container hosts a complete FL-Net *Site* environment. Each independent *Site* has its own local database, identity service, execution runtime, and connects to a common FL-Net *Platform*. Each *Site* ran as an isolated environment with its own data partition. This reproduced the *one-connection-per-Site* topology of a real deployment while allowing the entire federation to be instantiated, torn down, and re-run deterministically from a single host. The *Platform* recorded *Workflow* states, query histories, aggregation metadata, hyperparameters, and artifact versions across all experiments so that any run can be reconstructed in full from its stored provenance.

All experiments were performed on a Linux server equipped with one AMD EPYC 7401P 24-core processor with simultaneous multithreading enabled, providing 48 logical CPUs, and 188.8 GiB RAM. The system runs Ubuntu 24.04.4 LTS with Linux kernel 6.8.0-110-generic, Docker Engine 29.4.2, Docker Compose v5.1.3, and the overlay2 storage driver.

## S3.1 Data preparation and partitioning

The same harmonization and preprocessing pipeline was used for all conditions. First, the raw US-130 datasets were converted into the broad *Schema* used by the *Tool*.
Columns for identification, redundant age fields, missing columns, and columns with low information content were excluded from the modeling. Age classes were mapped to numerical means. Diagnosis codes were grouped into broad clinical categories. The medication columns were binarized to indicate whether a medication was prescribed or changed. The remaining categorical variables were one-hot encoded and the remaining numeric missing values were imputed using the median of the respective column.
Training and evaluation partitions were defined prior to model fitting. For federated experiments, the training pool was split into *Site-specific* client datasets. Each client then performed a stratified training/validation split with a 20% validation proportion and a random seed of 42. For centralized evaluation, the training dataset was a filtered dataset corresponding to a specific experiment and number of clinics.
All centralized runs were evaluated using the same held-out clinics corresponding to the synthetic clinics 120–130.

## S3.2 Trainable Tool

The model was implemented as an MLPClassifier from the scikit-learn library. This architecture was chosen because it utilizes a simple, well-established neural network for processing. The standard architecture consists of two hidden layers with 12 and eight neurons, respectively. These layers use the ReLU activation function, while the output layer is based on binary logistic classification. Training was performed using the Adam optimizer with a learning rate of 0.001, an L2 regularization strength of 0.0001, a batch size of 10, and a fixed initialization value of 42 (random seed).
In the centralized approach, training was always performed with 10 iterations. In the federated approach, training was also performed with 10 federation rounds and one local epoch per round. Consequently, both the centralized and federated training had the same number of iterations.
The federated model was optimized through FedAvg (*3*).

## S3.3 Results

Across an increasing number of clients, federated training has been shown to enhance client-level AUROC and AUPRC when compared to isolated local models (fig. S3.1). The most substantial gains were observed in the range of 25 to 50 clients. The per-client boxplots (S.3.3.1-S3.3.5) demonstrate that the benefit was not uniform, with some clients experiencing substantial gains, while others exhibited minimal change or occasional degradation. Typically, it exhibited an adverse response during client-level validation but a favorable one in the context of unseen data relative to central training. This approach facilitates the discrete reporting of discrimination and calibration outcomes.

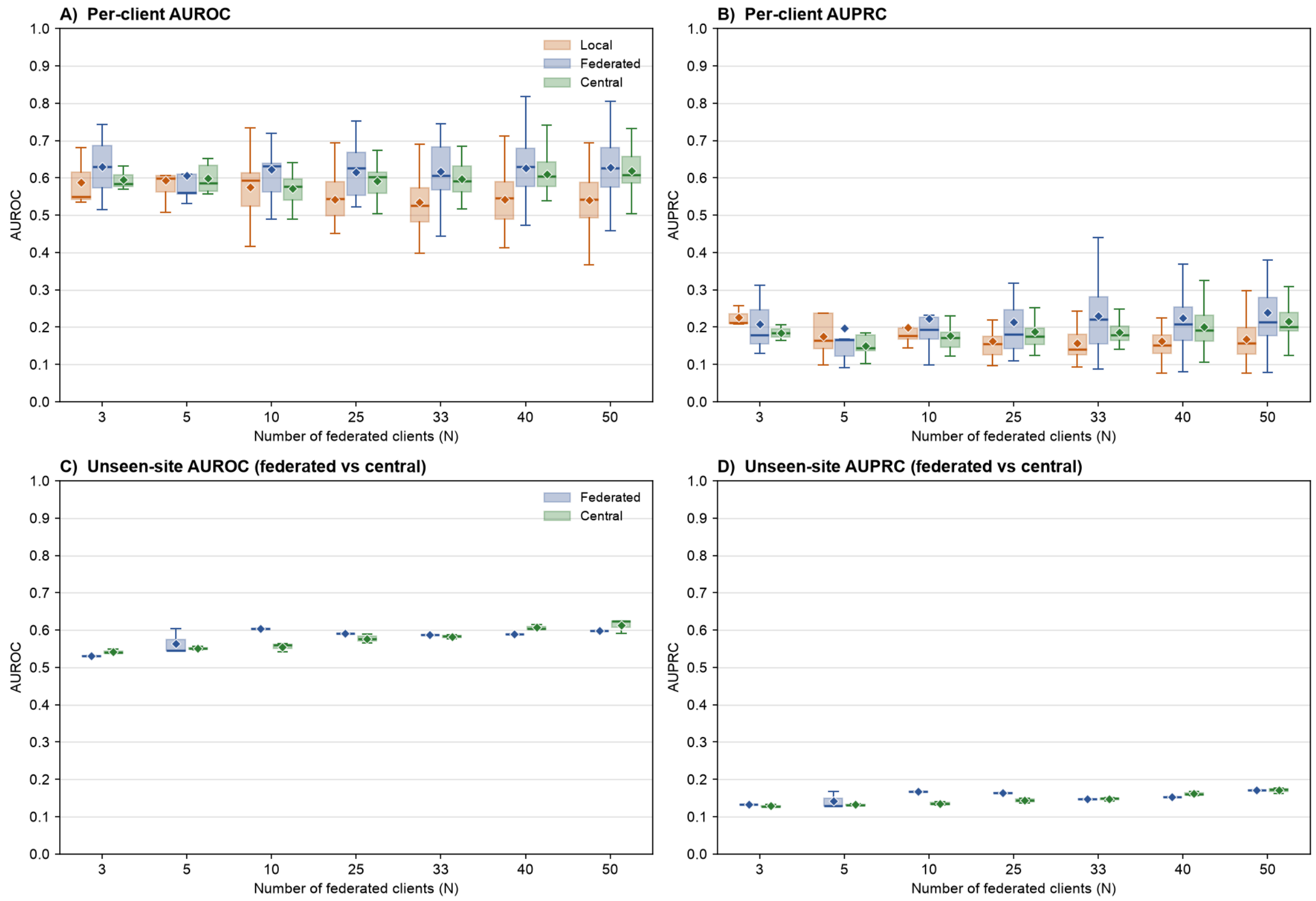

**Supplementary Figure S3.1 - Federated model evaluation:** Federated learning improves discrimination among participating clients compared with isolated local training. Unseen-clinic performance is compared against a fair central baseline trained on the same first N client data regime. A) Per-client AUROC on each participating clinic's held-out 20% validation split. The central model is trained using the same first N training *Sites* as the federation, with each clinic split into an 80/20 training/validation set, as in the *FL-Tool*. The boxes show the distribution across the N participating clinics: local clinic-only models, the federated model evaluated per clinic, and the fair central model evaluated per clinic. For the central condition, each clinic value is averaged across three central runs before plotting. B) Per-client AUPRC on the same participating-client validation splits, plotted with the same box-and-mean-marker convention as panel A). C) AUROC on the fixed unseen-clinic evaluation set (last 30 clinics) for the final federated and central models. D) AUPRC on the same unseen-clinic evaluation set, using the same three-run/model comparison as panel C). The unseen evaluation set contained 20,351 encounters with a readmission prevalence of 11.2%.

Federated scalability was assessed across federation sizes ranging from 3 to 50 clients over 10 training rounds and three repeated runs (fig. S3.2). Although absolute round duration increased with the number of participating clients, the amortized runtime per client decreased and remained low at larger federation sizes. Communication accounted for a relatively stable fraction of the total round time, while round-level measurements showed consistent behavior across repetitions and training rounds. Together, these results indicate that FL-Net scales to larger federations without a proportional increase in per-client execution cost.

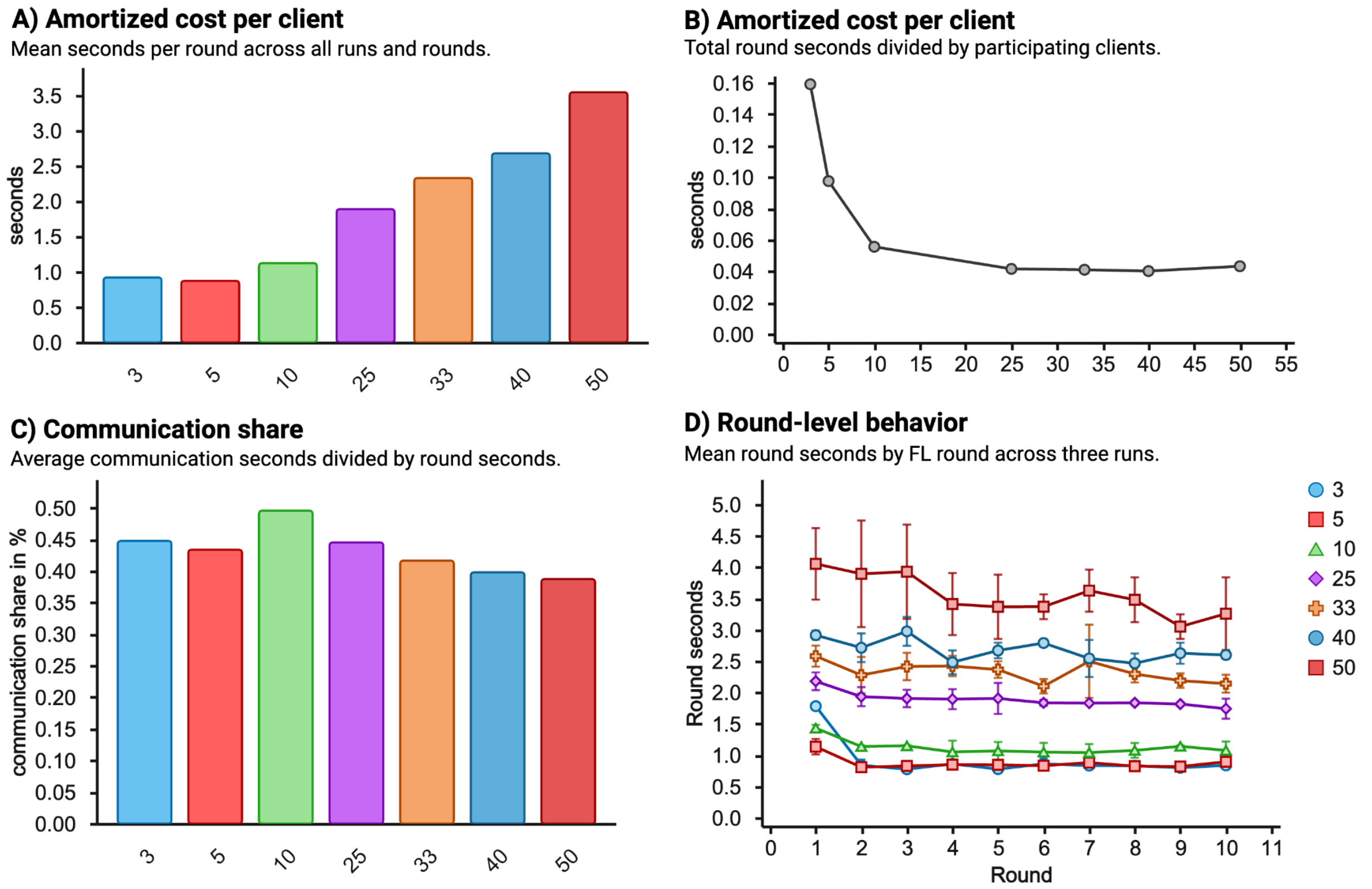

**Supplementary Figure S3.2 - Federated scalability evaluation:** The figure summarizes the runtime and communication characteristics of the federated neural network across different federation sizes. The values shown are averages across three repetitions and 10 federated training rounds per setting. A) shows the absolute average round duration. B) shows the same round duration normalized per participating client. C) quantifies the proportion of communication relative to the total round duration. D) shows the mean round duration across the 10 training rounds for each federation size. The number of clients examined were 3, 5, 10, 25, 33, 40, and 50.

## S3.3.1 - Three Sites Evaluation

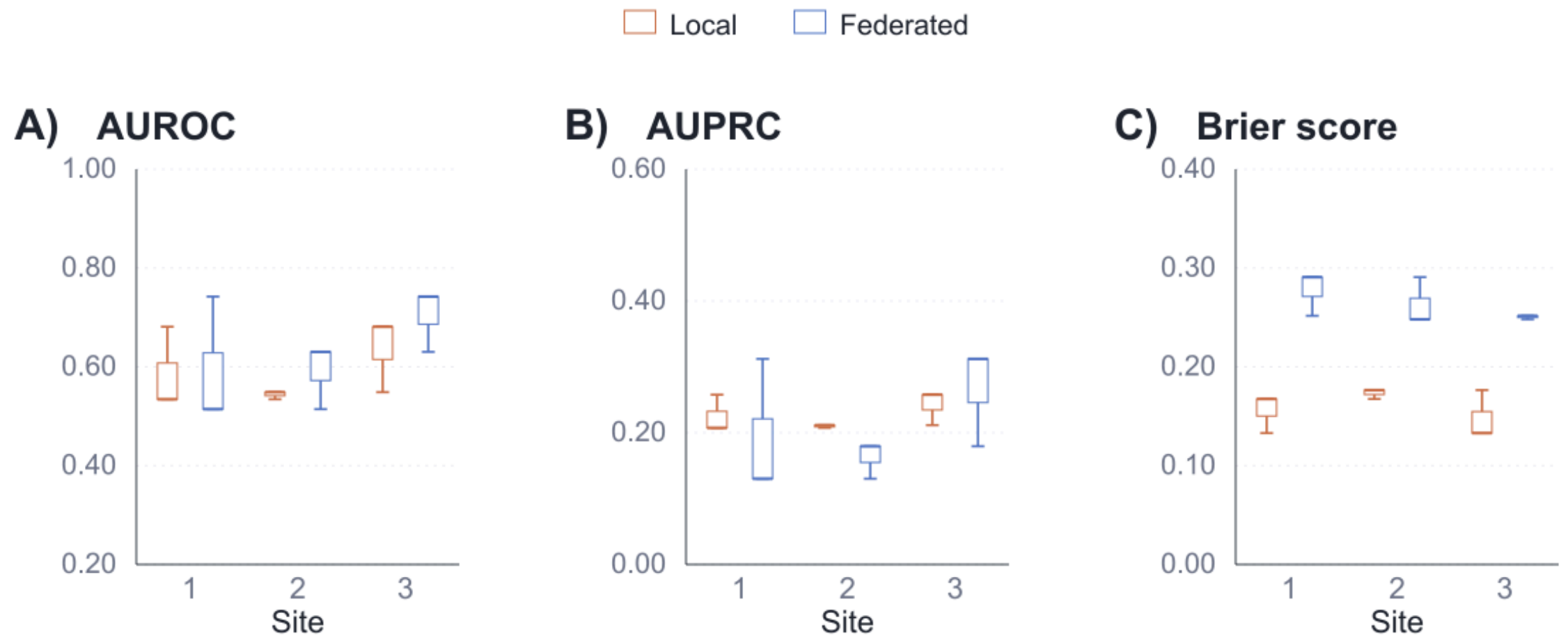

**Supplementary Figure S3.3 - 3 Site's Evaluation:** Per-client validation performance for the 3-client federation. Boxplots compare isolated local models with the federated/global model for AUROC, AUPRC and Brier score across available runs. The Brier score evaluates the accuracy of probabilistic predictions by measuring the mean squared difference between predicted probabilities and observed outcomes (*4*). FL improved mean AUROC but reduced AUPRC and worsened client-level Brier score, indicating limited and heterogeneous benefit at the smallest federation size.

The implementation of federated training resulted in an enhancement of client-level AUROC from 0.588 to 0.629 (fig. S3.3) However, this development was accompanied by a decline in AUPRC from 0.226 to 0.207, and a deterioration in Brier score from 0.159 to 0.264. In the context of unseen clinics, the federated AUROC was 0.538, in contrast to 0.659 for central training, suggesting limited generalization at this modest federation size.

## S3.3.2 - Five Sites Evaluation

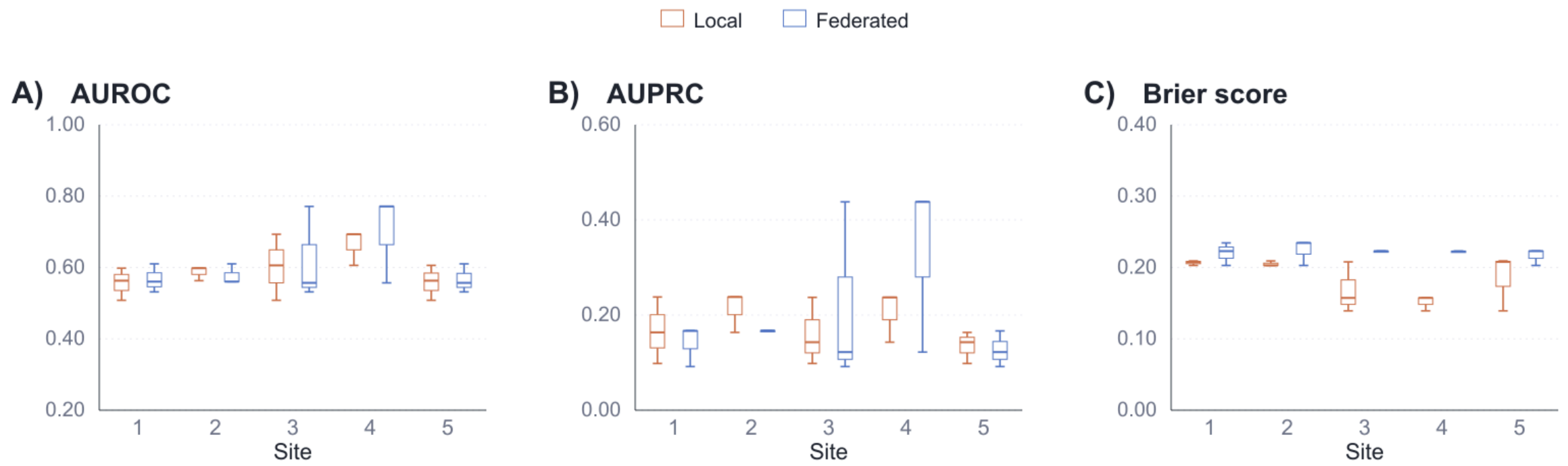

**Supplementary Figure S3.4 - 5 Sites Evaluation:** Per-client validation performance for the 5-client federation. Each client index is represented by paired local and federated boxplots for AUROC, AUPRC, and Brier score. Federated training produced modest gains in AUROC and AUPRC, while the Brier score increased, suggesting improved discrimination but weaker client-level calibration.

With 5 clients, federated training produced a modest AUROC gain (0.594 to 0.606) and improved AUPRC (0.176 to 0.197), while Brier score worsened (fig. S3.4). The unseen federated AUROC remained below central training (0.543 vs 0.663).

## S3.3.3 - 10 Sites Evaluation

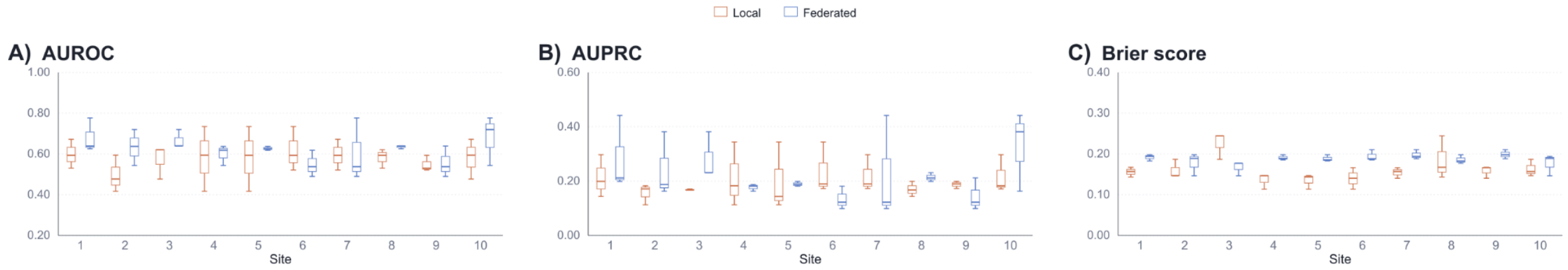

**Supplementary Figure S3.5 - 10 Site's Evaluation:** Per-client validation performance for the 10-client federation. Box plots summarize local and federated model performance across runs for each client index. Federation improved the mean AUROC and AUPRC across clients, with visible heterogeneity between clients, while the Brier score remained less favorable than the local baseline.

With 10 clients, the federated model improved AUROC from 0.575 to 0.623 and AUPRC from 0.198 to 0.222 (fig. S3.5). However, unseen AUROC remained lower than the central reference (0.543 vs 0.654), suggesting improved client-level performance but incomplete recovery of pooled-data generalization.

## S3.3.4 - 25 Sites Evaluation

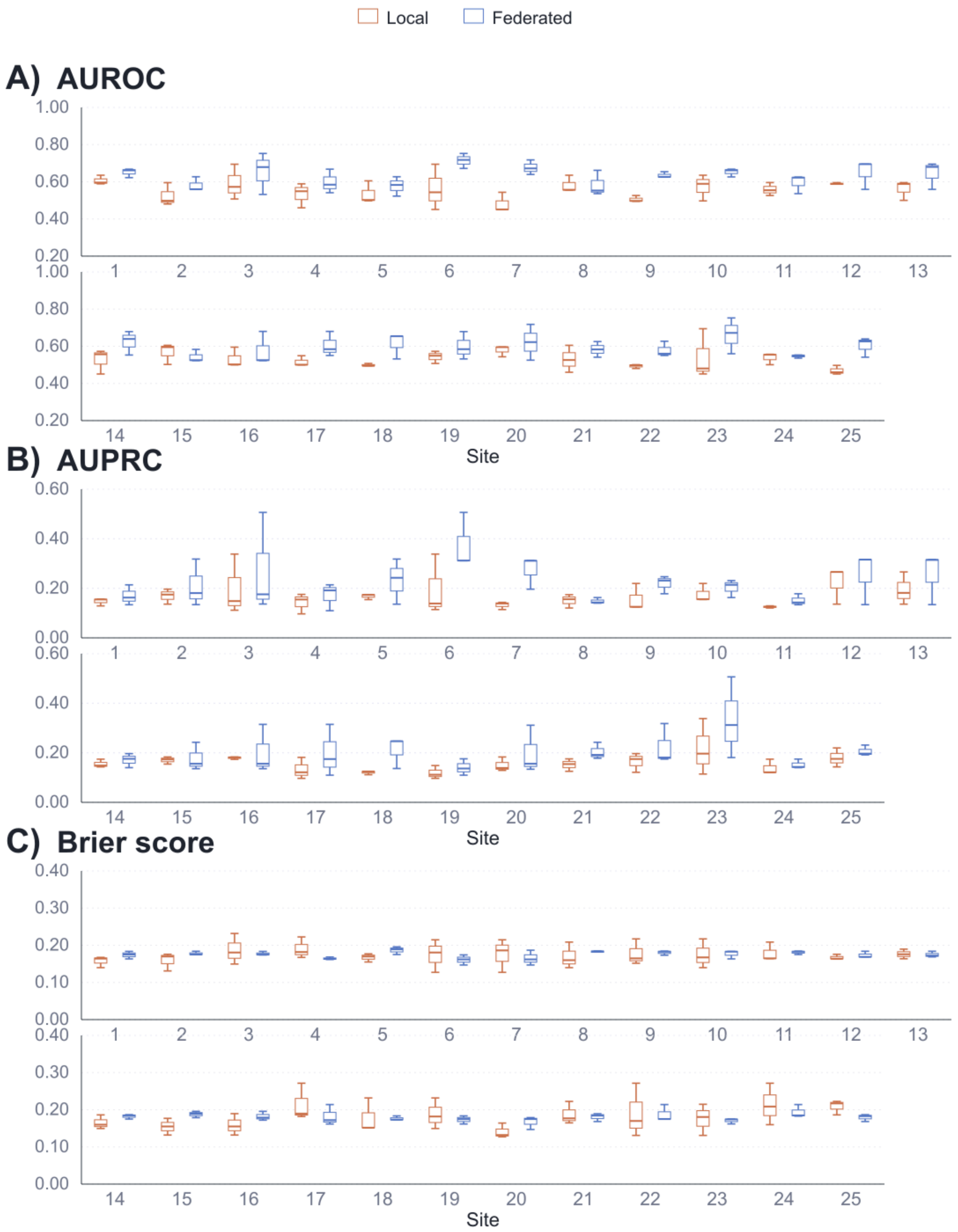

**Supplementary Figure S3.6 - 25 Site's Evaluation:** Per-client validation performance for the 25-client federation. Paired boxplots show isolated local and federated/global performance for AUROC, AUPRC, and Brier score. Federation improved AUROC and AUPRC for most client validations, indicating broad discrimination gains, whereas the Brier score showed little improvement on average.

With 25 clients, federated training showed stronger client-level gains, improving AUROC from 0.542 to 0.615 and AUPRC from 0.162 to 0.214 (fig. S3.6). Improvements were observed in 80% of client validations for both metrics. The unseen federated AUROC was 0.549 compared with 0.651 centrally.

## S3.3.5 - 33 Sites Evaluation

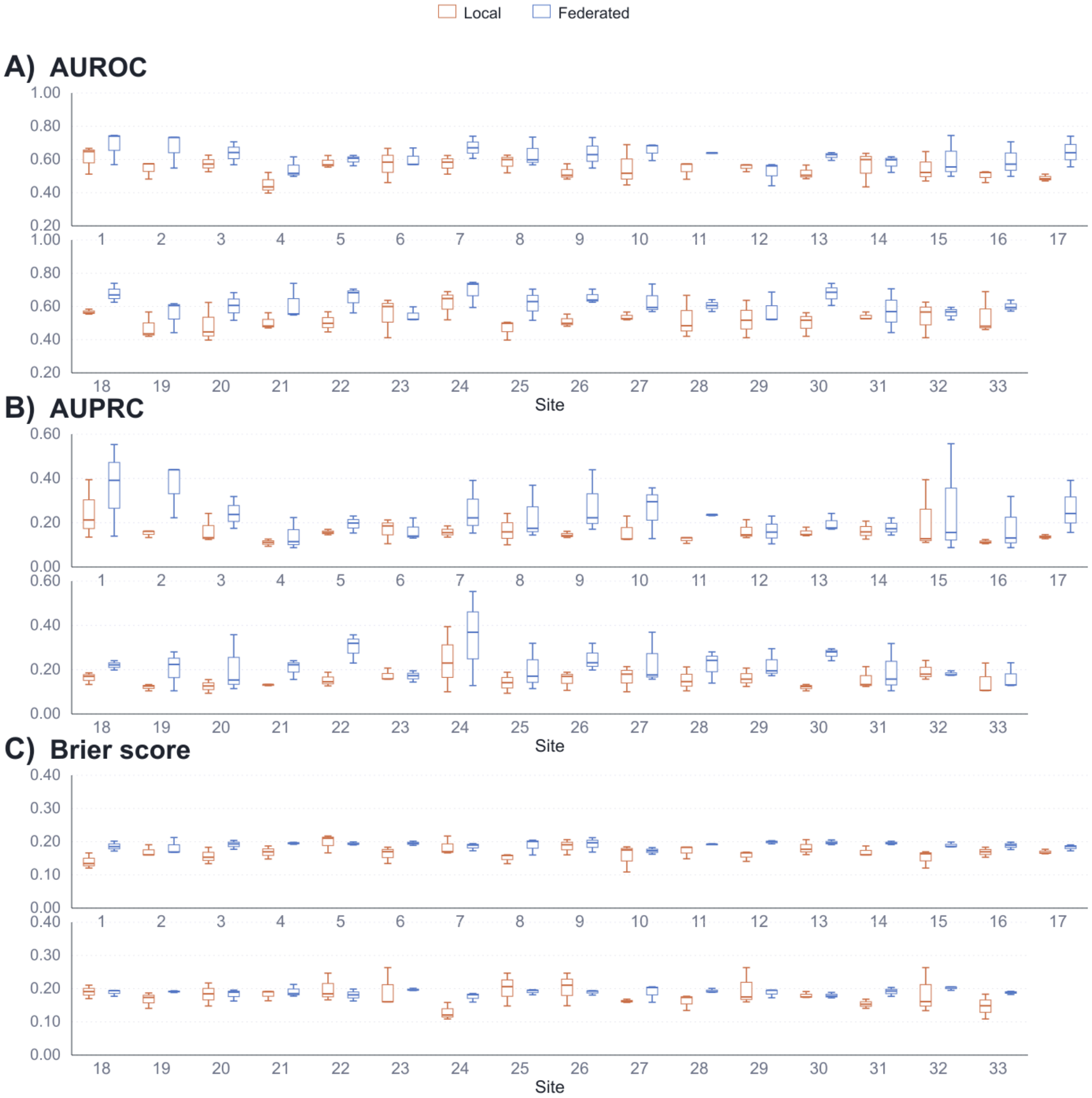

**Supplementary Figure S3.7 - 33 Site's Evaluation:** Per-client validation performance for the 33-client federation. Local and federated/global models are compared on a per-client basis for AUROC, AUPRC, and Brier score. Federated training yielded substantial improvements in AUROC and AUPRC across many clients, but the client-level Brier score generally increased, suggesting a trade-off between discrimination and calibration.

With 33 clients, federated training improved AUROC from 0.535 to 0.617 and AUPRC from 0.158 to 0.230 (fig. S3.7). *Site-level* gains were broad but heterogeneous. The unseen federated AUROC was 0.550, remaining below the central reference of 0.656.

## S3.3.6 - 40 Sites Evaluation

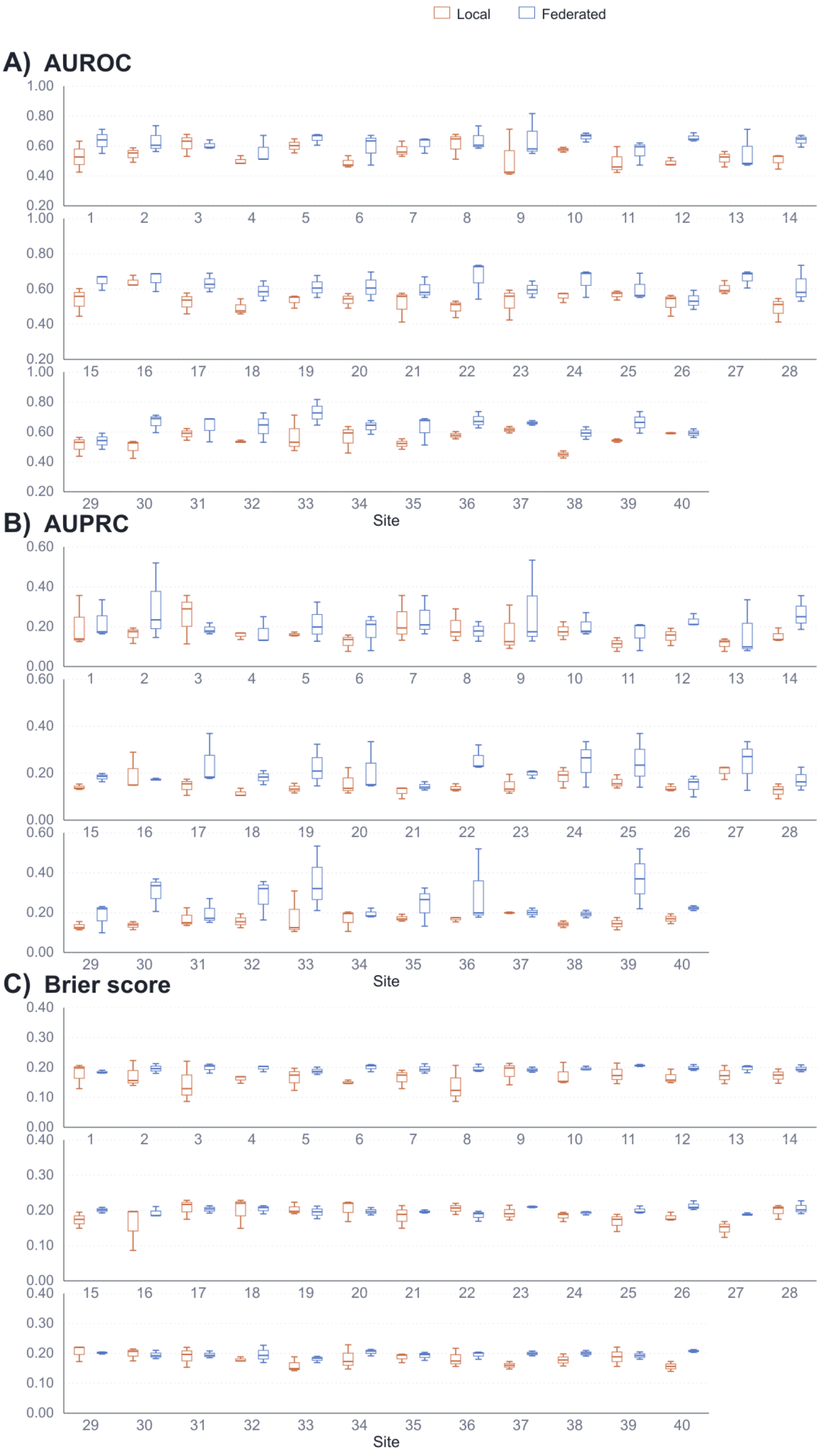

**Supplementary Figure S3.8 - 40 Site's Evaluation:** Per-client validation performance for the 40-client federation. Each panel compares client-wise local performance with federated/global performance for AUROC, AUPRC, and the Brier score. Federation yielded consistent improvements

in discrimination, particularly for AUPRC, while the Brier score remained inconsistent and often favored the local models.

With 40 clients, federated training improved AUROC from 0.541 to 0.624 and AUPRC from 0.160 to 0.222 (fig. S3.8). This setting yielded the highest unseen federated AUROC among the evaluated groups (0.562), although it remained below central training (0.656).

### S3.3.7 - 50 Sites Evaluation

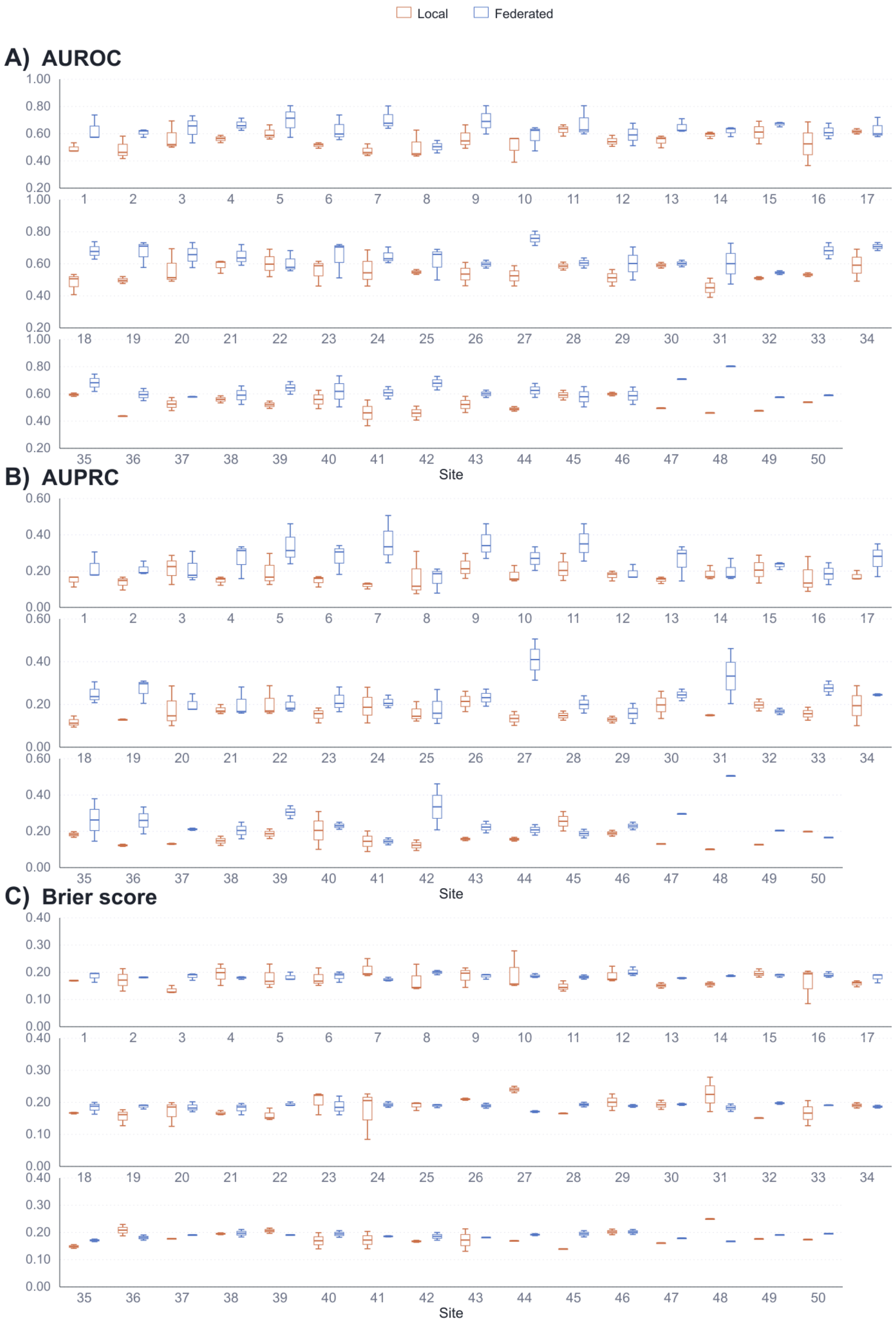

**Supplementary Figure S3.9 - 50 Site's Evaluation:** This is the per-client validation performance for the 50-client federation. Boxplots compare isolated local models with the federated/global model across client indices and runs. Although Brier score did not improve consistently, this setting showed the largest client-level AUROC and AUPRC gains, indicating that federation most strongly benefited discrimination rather than calibration.

With 50 clients, federated training produced the largest client-level gains: AUROC increased from 0.542 to 0.635 and AUPRC from 0.169 to 0.245 (fig. S3.9). Most client validations improved for discrimination metrics. On unseen clinics, federated AUROC was 0.554 compared with 0.654 centrally, showing strong client-level benefit but persistent central-model superiority for unseen discrimination.

# Supplementary 4: Implementation details

FL-Net is implemented as two multi-service components, the FL-Net *Platform* and the FL-Net *Site*, which together provide the infrastructure for federated data discovery and FL (fig. S4.1). To support *Tool* development, we provide an FL-Net *Tool* template written in Python that allows developers to create both *FL-Tools* and *ETL-Tools* for use within the system. While an FL-Net *Tool-API* is also available for direct implementation, use of the template is recommended. Further documentation is available under https://federated-learning.net/documentation, with specific documentation on the technical architecture available at https://federated-learning.net/documentation/docs/intro/architecture/welcome.

## Shared Services

Several services are used by both the FL-Net *Platform* and the FL-Net *Site*, with each *Site* and the *Platform* deploying their own instance of each service.

The Orchestration API (Orch API) handles the lifecycle of both ETL and FL-related *Tools*. It is implemented using the Quarkus framework in Java and exposes an HTTP server that is accessed exclusively by the Local or Global Learning APIs. Persistence is provided by an additionally deployed PostgreSQL instance (Chainguard PostgreSQL image). On the *Platform* the Orchestration API is currently used for centralized *Tool* execution as described in PoSyMed (*5*) and will be used for on *Platform* deployment of the aggregating component in FL runs in a future version.

The Controller manages all data exchange occurring during an FL *Project* run, between the running *FL-Tools*. It is based on the FeatureCloud Controller (*6*), but has been adapted for the FL-Net *Tool* API. The Controller is implemented in Go and is intended for secure FL message communication handling including optional privacy enhancing techniques. It deploys an HTTP server implementing the *Tool* API allowing *FL-Tools* to use it to receive and send messages. The Controller acts as a TCP Client to the Relay Server to forward and receive messages for the *Tool*. As part of the adaptation of the Controller, an in-memory message store for incoming messages was added to the Controller and a message queue for outgoing messages was added to both the Controller and the Relay Server described as part of the *Platform* services. The Controller protects each FL *Project* run's data sending and retrieval endpoints by a dedicated API key that is communicated only to the associated FL-*Tool*, preventing concurrently running *FL-Tools* from accessing data belonging to other runs. The Controller is deployed on the *Platform* to support deploying the *Project's* aggregation component in a future release.

Keycloak provides authentication and authorization. Each FL-Net *Site* deploys its own Keycloak instance, and the FL-Net *Platform* maintains a separate instance of its own. In addition, every FL-Net *Site* must hold an account within the FL-Net *Platform's* Keycloak instance, which protects the *Platform* against unauthorized or malicious FL-Net *Sites*. As with the Orch API, persistence is handled by an additionally deployed PostgreSQL instance (Chainguard PostgreSQL image).

## FL-Net Platform Services

The frontend is used by global *Researchers* to perform *DDQ*, retrieve statistics, submit *FL-Requests*, manage *Schemas* and manage and use models and is additionally used by the *Platform-Admin* to monitor service health and the *Auditor* to review and certificate *Tools*. It is implemented in Angular and containerized using the Chainguard nginx image to serve the compiled application.
The Global Learning API coordinates and manages the FL *Project* run lifecycle, authenticates and coordinates FL-Net *Sites*, orchestrates FL *Project* runs, and handles *Tool* storage, including secure *Tool* building and auditing, as well as model storage and the automatic generation of *Inference-Tools* from FL *Project* runs. It is implemented using the Quarkus framework in Java and exposes a HTTP API which is accessed via the browser by a *Researcher* for *DDQ*, statistics retrieval, and *FL-Requests*. Each FL-Net client also accesses this server to receive and respond to federated requests and to download *Tools* for local execution. Lastly, this server is accessed by other services of the *Platform*. Persistence is provided by an additionally deployed PostgreSQL instance with pgvector support. By incorporating pgvector, FL-Net retains full compatibility with all Retrieval-Augmented Generation and Large Language Model capabilities previously introduced in PoSyMed (*5*).
The Data Modeler API manages *Schemas, Ontologies, DataTypes*, and their associated embeddings. The embeddings help search through the millions of $N_O$ which typically occur in biomedical ontologies. It is implemented using the Quarkus framework in Java and exposes an RESTful HTTP server that is accessed via the browser by global *Researchers* for *Schema* and *DataType* management, by FL-Net *Sites* for *Schema* retrieval in cohort creation, and by a global users for more extensive *Schema*, *Ontology*, and *DataType* management. Persistence is provided by an additionally deployed Neo4j instance.
The Relay Server handles communication between the local Controllers to ensure the secure exchange of e.g. model parameters during FL *Project* runs. It is based on the FeatureCloud Relay Server *(6)*, adapted for FL-Net's FL orchestration requirements, and is implemented in Go. It deploys an HTTP server for FL *Project* run management, used exclusively by the Global Learning API and isolated to the *Platform's* own docker network, alongside a TCP server with custom framing used by FL-Net *Sites* exposed on an additional port. A separate chainguard NGINX image based container terminates TLS and acts as a reverse proxy for the backend services.

## FL-Net Site Services

The frontend is used by the *Site-Owner* to control *DDQ*, statistics, and learning,check auditing logs of any federated secured access, to administer the *Site*'s (patient) data and to monitor service health. As with the *Platform* frontend, it is implemented in Angular and containerized using the Chainguard nginx image.
The Local Learning API manages local data storage, *Tool-assisted* ETL import, *DDQ*, statistics and learning access and access control, and maintains extensive auditing of changes to patient data as well as all federated access, including federated queries, statistics, and learning requests and runs. It is implemented using the Quarkus framework in Java and exposes a HTTP API which is accessed exclusively by the client's *Site-Owner*, while the Local Learning API itself acts as an authenticated WebSocket client to the FL-Net *Platform*, through which it receives and responds to federated queries, statistics requests, and learning requests, and synchronizes FL *Project* run orchestration. Persistence is provided by an additionally deployed PostgreSQL instance (Chainguard image).
A separate chainguard NGINX image based container terminates TLS and acts as a reverse proxy for the backend services.

## Communication and Encryption of FL messages

Communication between *FL-Tools* is mediated by the Controller service, which relays messages to their intended recipients via the Relay Server. All data exchanged between clients, or between a client and the aggregator during training, is additionally encrypted on top of standard TLS using ECIES-style encryption over X25519, which replaces the NIST P-224 curve previously used by FeatureCloud while retaining the same underlying encryption technique. The shared secret is derived via ECDH, combining the recipient's per-FL-run ephemeral public key with a per-message ephemeral private key generated by the sender. This additional encryption layer does not apply to broadcasts from the aggregator to the clients, which rely on TLS encryption alone for performance reasons: since the scheme requires per-recipient encryption, applying it to broadcasts would force the aggregator to send individual messages to each client rather than a single message distributed to all clients by the Relay Server.
Authentication proceeds on two levels. Mutual authentication between the Controller and Relay Server is achieved via mTLS at the transport level. On top of this, application-layer authentication of clients toward the Relay Server is implemented through API keys, generated by the Relay Server per FL *Project* run and distributed via the Global and Local Learning APIs to each Controller, which transmits its key as the first message upon connection; connections presenting keys that are invalid for the relevant FL *Project* run are rejected.

## Maintenance and Security

To ensure ongoing security and maintainability, FL-Net undergoes regular scans for container and dependency vulnerabilities, as well as secret detection and static code analysis.
*Tools* deployed on the FL-Net *Platform* are constructed directly from their corresponding Git repositories. The *Building-Pipeline* utilizes the repository Dockerfile to execute vulnerability and malware scans. Upon execution, the *Building-Pipeline* binds the security scan outputs, the generated Docker image, and the functional testing dataset to the specific Git commit hash, as PoSyMed suggested (*5*). PosyMed's testing *Building-Pipeline* has been expanded to simulate the functionality of multiple clients, thereby also supporting FL unit tests (*5*). For this purpose, multiple data sets can be assigned or generated dynamically.
During the *Tool* audit, an authenticated *Auditor* reviews the *Tool* version using the source code pinned to the commit hash along with the associated *Building-Pipeline* outputs. *Tools* are awarded up to five auditing stars, where each star denotes formal approval by an individual *Auditor*. This auditing protocol establishes verifiability and increases trust in reviewed *Tool* versions.

# Deployment

Two deployment repositories are provided. One is for the FL-Net *Site*. The other is for the FL-Net *Platform*. This enables self-hosted deployment of FL-Net. Both deployment folders are based on Docker Compose and include a Python script that generates initial secrets and guides users through configuration persisted via environment files; this script has no dependencies beyond the Python 3 (≥3.6) standard library. Running the FL-Net *Site* requires only Docker and Docker Compose, whereas the FL-Net *Platform* additionally requires an SSL certificate and a registered domain; the latter requirement is optional for the FL-Net *Site*. The FL-Net *Site* deployment folder further includes a backup shell script that consolidates all persisted data and relevant generated secrets, such as database passwords, into a single timestamped folder. Further documentation on the backup process is available at https://federated-learning.net/documentation/docs/deployment/backup-client.

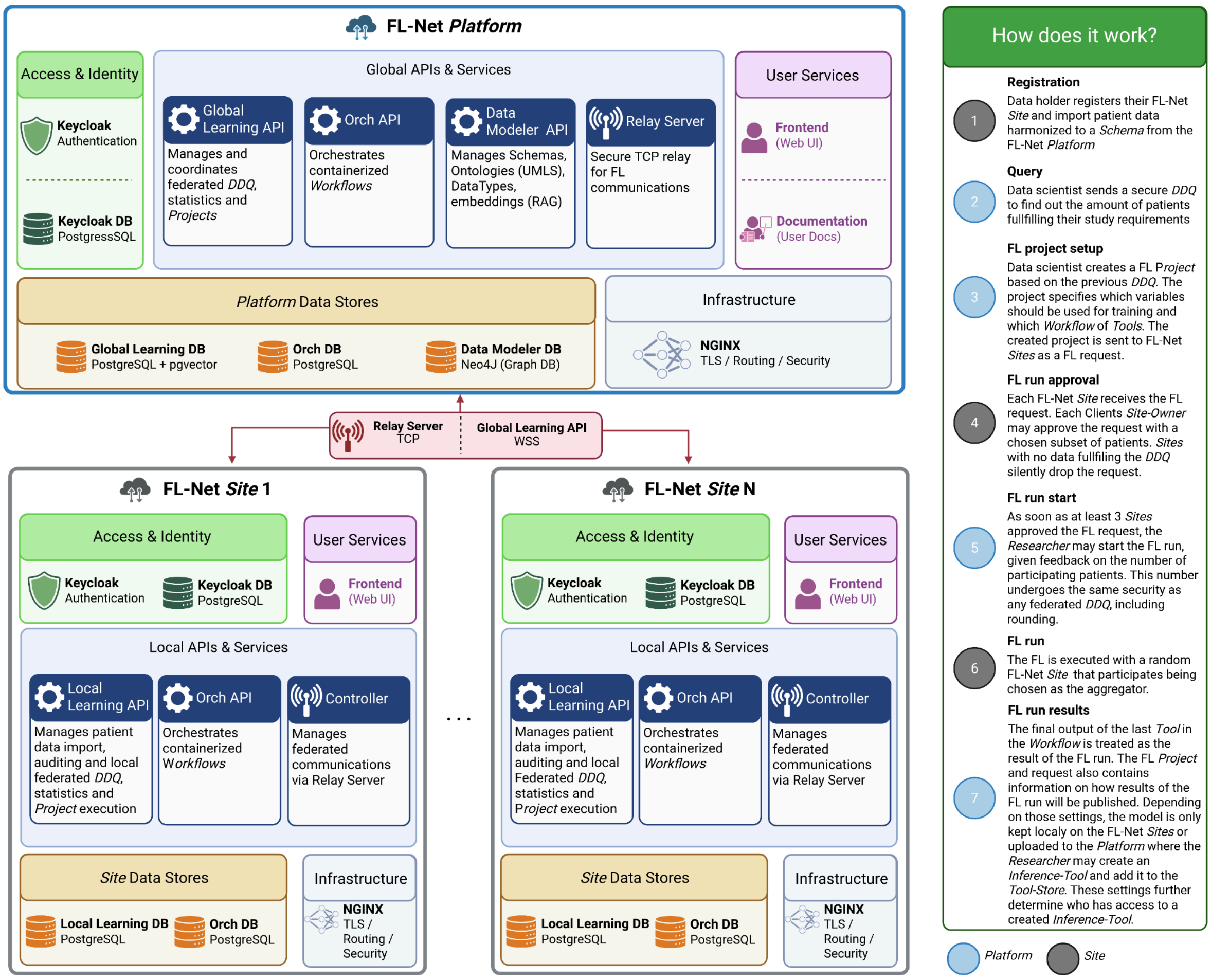

**Supplementary Figure S4.1 - System architecture and operational flow of FL-Net:** FL-Net follows a star shaped network architecture, where a central *Platform* coordinates multiple distributed *Sites* without requiring direct access to local, patient-level data. Both the FL-Net *Platform* and the FL-Net *Site* are protected through access and identity management by their own Keycloak instance. They also both expose APIs and services via HTTP endpoints accessed via their relative NGINX for routing, TLS and additional security. FL *Project* run communication is handled via Controller services per FL-Net *Site* relaying messages via the Relay Server deployed on the FL-Net *Platform*. The FL-Net *Platform* uses PostgreSQL with pgvector to also store embeddings, and Neo4j to store the graph structure of *Schemas*, $N_O$ and $N_{DT}$. The FL-Net *Site* only requires vanilla PostgreSQL data storage. Additionally, the FL-Net *Platform* provides the documentation as an extra service. The typical

operational flow in FL-Net starts with data holders deploying their own FL-Net *Site* and importing data according to an existing *Schema* or creating a *Schema* describing their data. In the next step, a *Researcher* runs a *DDQ* to see if a sufficient amount of patient data could be used for FL. If they are satisfied with the amount, they create a FL *Project*. This contains the information of the *DDQ* and which variables should be used for training, the *Workflow* to be executed, and whether the resulting model should be published. When the FL *Project* is created, it is packed into a *FL-Request* sent to all *Sites*. If the DDQ was previously answered with a non zero count it is repeated. If the repeated DDQ yielded a non zero count, the Data holder can review the request. The request includes the *Project* configuration, selected *Tools*, and their settings, as well as certifying audit reports provided by specific *Auditors*. For example, a *Tool* may have up to five verified stars indicating that it has been reviewed and approved by five *Auditors*. Data holders can then allow the request for a subset of patients of their choosing. After at least 3 *Sites* (configurable on Platform level) approved the request, the *Researchers* may start the FL *Project*. Any *Site* that didn't yet approve the relevant request cannot join anymore now. Each participating *Site* is informed of the start of the *Project*, pulling the relevant *Tool* and exporting the relevant data into their locally executed *Tool*. Finally, the aggregator of this FL *Project* run, running on a random *Site*, uploads the resulting model to the FL-Net *Platforms* model store with appropriate access settings depending on FL *Project* settings. In case of maximum model privacy, the final model is only distributed to the FL-Net *Sites* but not to the FL-Net *Platform*.

# Supplementary 5: Other federated learning frameworks

Several existing FL frameworks cover key components of the infrastructure required for multicenter clinical research but differ significantly in their scope of application. To enable a structured comparison, we evaluate each framework based on the five defined research requirements. To avoid the confusion of conflicting framework vocabularies, we reuse the terminology established in FL-Net. Note that this is purely a conceptual mapping and does not mean components like FL-Net *Tools* are cross-compatible. This neither means that the requirements are not met if any framework solves the concepts behind the vocabulary in a different way. For instance, the usage of a sequential workflow system is as valid as a DAG-based workflow system. Furthermore, please note that this is not meant as a ranking of frameworks but merely to distinguish the end-to-end capabilities as defined by the authors via the requirements backed by the literature. We acknowledge that each framework was created for its own specific purpose and completely fulfills its intended purpose.

**RE1 - Semantic and syntactic data harmonization:** This requirement involves converting heterogeneous local data into a common representation based on shared concepts and validation rules enforcing quality control. FL-Net implements this through $N_O$, $N_{DT}$, *Schemas*, and validated ETL processes that map and validate local data. Full support requires an integrated semantic and syntactic harmonization layer that firstly can be shared by all participating *Sites* and secondly enforces validation rules ensuring quality control. Partial support is met if a framework provides built-in preprocessing, mapping, data loaders, or dataset-specific transformations without a common semantic model and validation. The requirement is not met if data preparation is left entirely to individual applications or users, only providing an entrypoint for harmonization logic but no functionality itself.

**RE2 - Secure, Authenticated, and Auditable Access:** This requirement mandates authenticated and encrypted remote access for *Site-Owners*, institutional control over all patient data disclosed through federated access, and auditing of all interactions with patient data. FL-Net provides authenticated and encrypted *Site* access and fine-grained control over federated data disclosure for *Site-Owners*. *Tools* are securely executed in isolation and use encrypted communication. All interactions with patient data are logged for auditing.
This requirement is met if (i) technical safeguards, (ii) institutional disclosure control and (iii) auditing of any federated access on at least patient level are all fulfilled.
(i) For technical safeguards, any access must be authenticated. All external communication, including the federated learning channel, must be encrypted. Code on private data must be executed in isolated environments.
(ii) For institutional disclosure control, the *Site-Owner* needs to be able to authorize and control any external access, irrespective of whether it's for discovery or learning. Any learning should be reviewable.
(iii) Lastly, for auditing, the FL framework must provide auditing that logs any external interaction with patient data.
Partial support is met if all technical safeguards (i) are supported, the requirement is not met otherwise. Please note that requiring manual data addition to a specific workflow is not considered disclosure control by this criteria.

**RE3 - Reusable and Versioned Federated Workflows:** This requirement addresses the question of whether FL can be defined as persistent, reusable, and versioned, rather than being recreated for each study. FL-Net treats *Workflows* and *Tools* as reusable, versioned, publishable store objects with defined inputs, outputs and hyperparameters. Full support is met when an executable version of *Workflow* and *Tool* objects including both their configuration options and executable components is persistently stored and can be reused. If no workflow system exists but the requirement is met for *Tools*, partial fulfillment is achieved. Furthermore, any data scientist access based purely on notebooks only earns partial fulfillment, as these are reusable, but versioning is then optional and usually done externally, e.g. via git versioning. The framework may use external services such as an external Docker registry, as long as these are integrated into the usage of the framework, and still earn up to complete fulfillment.

**RE4 - Identification of cohorts while preserving patient privacy:** This requirement concerns determining whether *Sites* have sufficient and appropriate data prior to training, without disclosing patient-level data records. FL-Net performs ontology-based queries locally and returns only permission-controlled, aggregated counts. Full support is met when a dedicated, *cross-Site* data feasibility query mechanism is available prior to training that protects individual data records through aggregation or equivalent disclosure controls but offers semantic exploration. This mechanism must encapsulate data discovery over multiple *Sites* simultaneously, any mechanism that is on a *per-Site* basis only gives partial support. Partial support also applies when distributed aggregated analyses can technically be used for feasibility assessment, but no dedicated semantic layer for *cross-Site* identification is available. If feasibility is purely based on assumptions of data formatting, this criterion is not met.

**RE5 - Traceability of Approvals, Versions, Participation, and Results:** This requirement addresses whether the entire execution context of a federated study remains traceable. The resulting model must be automatically linked with traceable information about the *Tools*, hyperparameters, and metadata of the data used for its creation to all parties authorized to access the model. This traceability is necessary to enable reproducible and verifiable federated studies. In FL-Net, *Projects*, participating *Sites* and their approvals, *Workflow* and *Tool* versions, runs, metrics, and resulting models are recorded and linked to the corresponding execution. Full compliance is met if, for each federated execution, approvals of participating *Sites*, executed *Workflows* including *Tools* and their version, and resulting outcomes are permanently recorded and automatically linked. This is also fulfilled in the case of using notebooks to document the FL execution, as they inherently contain the complete trace and output. The requirement is partially met when traceability is given but requires manual linkage (e.g., through log inspections, etc.), or if the federated execution of the *Workflow* is traceable by participants, but the outputs lose linkage, e.g. by being available via download. The requirement is not met if the framework does not produce clear, linkable logs for at least *Tool*/*Workflow* versioning of the FL execution.
Relevant *Site* pseudonymization mechanisms that allow only the respective *Site* to link its own participations, while exposing only pseudonyms to other entities, may be used without impeding RE5. Furthermore, as not all federated consortia wish to publish outcomes, any system that limits the accessibility to the trace of an FL execution also earns fulfillment.

## Apheris

Apheris is a commercial framework for federated data networks in the life sciences and biopharmaceutical sectors. It executes models and *Workflows* on distributed proprietary data, particularly in the areas of drug discovery, protein structure prediction, and foundation model applications. Datasets are given a name and a description by their relative data custodians, with these then available to the FL creating *Researcher*. This means harmonization and data discovery is done manually, therefore not meeting RE1. The datasets however can be explored via statistics, therefore giving partial fulfillment of RE4. Data custodians may set asset policies, enabling disclosure control and ensuring RE2 is met (*7*, *8*). The FL execution management is done in a notebook style, where a *Researcher* creates and runs a notebook. This inherently has no direct versioning support, only partially fulfilling RE3, but fulfilling RE5 as the notebook contains the complete trace of the FL execution (*7*, *8*). In summary, Apheris does not fulfill RE1, partially fulfills RE3 and RE4 and fully supports RE2 and RE5 (Table 1), but it should be noted that it's a commercial, non open source solution focused on single consortia.

## CAFEIN

Developed within the CERN community, CAFEIN is a federated AI framework that enables collaborative ML across institutions and data spaces. Its main contribution is extending large-scale scientific computing framework to federated AI *Workflows*. However, compared to FL-Net, CAFEIN is primarily a general-purpose FL execution framework rather than a clinical semantic framework. The ontology-based structure is also not a central focus (*9*). RE1 is not fulfilled, while CAFEIN offers modular and extensible data loaders, these must be added by the *Sites* themselves and therefore coordinated within the FL execution. RE2 is partially met here as CAFEIN focuses on technical safeguards, but no specific disclosure control mechanism is implemented (*9*). *Tools* and *Workflows* are described in JSON format as settings like files. This does not implement any versioning system, and therefore RE3 is not met. However, any run FL is stored at the *Sites* and the *Platform* in attached storage volumes, storing local and global models, log files and TensorBoard files, linking them by their execution (*9*). Therefore, RE5 is fully met. Lastly, RE4 is not met as no specific data discovery is supported. In summary, CAFEIN does not fulfill RE1, RE3 and RE4, partially fulfills RE2 and fully fulfills RE5 (Table 1).

## DataSHIELD

DataSHIELD is an open-source framework for realtime, non-disclosive federated data analysis that enables multi-center statistical co-analysis without pooling or exposing individual-level records (*10*). Harmonization must be done as part of the *Workflow* itself. However, dsOMOP was developed and integrated into DataSHIELD, offering harmonization support for the OMOP data model (*11*, *12*). Therefore, RE1 is partially met, as harmonization exists but is limited to OMOP and the capabilities of the added package. RE4 is only partially met, as non-disclosive exploratory summary statistics are available, but cross domain data discovery is not supported. RE2 is only partially met. While e.g. role-based access control and server-side disclosure traps that evaluate every function output in real time to prevent the extraction of individual patient data exist, all of these are Server site protection mechanisms. However, *Site* specific disclosure control, especially on patient level, does not exist. RE3 is partially met as DataSHIELD functions specifically describe tools/algorithms and are versioned as DataSHIELD itself is versioned, but *Workflows* themselves are not directly versioned but handled as notebooks. However, as federated analysis is done via notebooks, these notebooks keep a complete trace of the FL execution, fulfilling RE5 (*10*). In summary, DataSHIELD partially fulfills RE1, RE2, RE3, RE4 and fully fulfills RE5 (Table 1) and only supports federated analysis using vetted functions but not full FL with community added algorithms.

## FATE

FATE is a specialized, open-source framework for collaborative and federated learning that meets data privacy requirements. It focuses on the production-ready implementation of collaborative ML *Workflows* for large-scale, distributed enterprise and institutional scenarios. Specifically, FATE addresses horizontal and vertical federated learning settings, as well as secure computational methods, which serve as the technical foundation for privacy-preserving model training (*13*). It fully supports the technical elements of RE2 related to data protection and secure data processing, but has no specific data disclosure mechanisms, therefore earning a partial fulfillment of RE2. Tools are handled by FATE itself as Algorithm components, with *Workflows* described as a directed acyclic graph of algorithms with additional information on participants and configuration. However the versioning of the algorithms is then handled externally via e.g. git. Therefore, RE3 is only partially met. Also, output of *Workflows* is handled via download, therefore only earning partial fulfillment of RE5. No specific harmonization support is given, resulting in no fulfillment of RE1. Data discovery is also only possible via federated statistics, earning only partial fulfillment of RE4 (*13*, *14*). In summary, FATE does not fulfill RE1, partially fulfills RE2, RE3 and RE4, and RE5 (Table 1).

## FeatureCloud

FeatureCloud (FC) is an integrated federated learning (FL) framework that enables users to assemble *Workflows* using containerized applications from its AI Store, executed locally via Controllers at each participating Site (*6*). However, FC requires *Sites* to manually join projects and provide pre-harmonized data, with preprocessing applications available via the AI Store and no specific framework support, earning no fulfillment of RE1. The absence of data discovery capabilities results in no fulfillment of RE4. Requirements RE2 is only partially met: although FC provides Docker isolation, restricted network access, secure communication, certification, and project-level execution tracking, local data governance relies on manual *Workflow* reviews, no specific disclosure control mechanism exists. Conversely, FC fully satisfies RE3 by allowing applications to be published, certified, structured into *Workflows*, and reused across projects, with documentation available per Tool for configuration options (*6*). However, RE5 is only partially fulfilled. While all involved *Sites*, including the *Site* that started the FL, store logs that can be linked to the executed Tools in the AI store of FC, this linkage must be done manually and is not done automatically by FeatureCloud. It should be noted that the FL communication in FL-Net is an adapted version of the FL communication used in FeatureCloud. In summary, FC does not fulfill RE1 and RE4, partially fulfills RE2 and RE5 and fully fulfills RE3 (Table 1).

## Fed-BioMed

Fed-BioMed is an open-source framework for FL in real-world biomedical and clinical contexts (*15*). Its key contribution is translating FL from experimental ML libraries into a transparent, trustworthy, hospital-ready execution environment (*15*). Notable features include clear separation of *Researcher* and non technical users, support for medical data custodians, and emphasis on governance, security, and usability (*15*). It fully satisfies RE2, as a Fed-BioMed *Site* has auditing available, specific disclosure control mechanisms and technical security (*15*). Fed-BioMed handles federated access via jupyter notebooks (*15*). Inherently, this only gives partial fulfillment of RE3 as versioning is then handled by other services such as git. However, RE5 is completely fulfilled as the notebook solution chosen by Fed-BioMed correctly collects the complete trace of an FL execution (*15*, *16*). RE1 is only partially met, as dataset loaders and local preprocessing do not constitute a full-fledged common validation layer (*16*). RE4 is only partially fulfilled, as only a free text description is available per dataset as well as tags, but no structured and in depth semantic and syntactic description (*16*). In summary, Fed-BioMed does partially fulfill RE1, RE3, and RE4 and fully fulfills RE2 and RE5 (Table 1).

## Flower

Flower is a universally applicable framework for converting existing machine learning workloads into federated applications for simulation and FL (*17*).
RE1 is not met, as data preparation remains application-specific and Flower does not provide for harmonization (*17*). As federated statistics is available, RE4 is partially fulfilled, requiring manual work across federated statistics results. RE2 is incomplete, as Flower documents TLS, Flower client authentication, and integrated secure aggregation, while audit logging is available as an enterprise feature and local approvals for the use of clinical data or disclosure policies at the cohort level not supported (*17*, *18*).

RE3 is partially fulfilled as the Flower Hub stores Tools as versioned, reusable objects, but *Workflows* are not stored (*17*, *18*). RE5 is completely fulfilled, as notebook style federated access can serve to record the complete stack of FL execution (*17*, *18*). Flower does not fulfill RE1, partially fulfills RE2, RE3, RE4 and fully fulfills RE5 (Table 1).

## MedPerf

MedPerf is an open framework for the federated evaluation and benchmarking of medical AI models (*19*). Instead of training federated models, MedPerf focuses on the standardized, privacy-preserving evaluation of registered models on distributed, external, and often previously unseen clinical datasets (*19*). A benchmark is registered as a regulated collection of containers for data preparation, reference models, and metrics and requires administrative approval, which supports reusable benchmark *Workflows* and registered artifacts. However, as the benchmark Tools are only for evaluation and not FL, requirements RE3 and RE5 are not met, as Tools defined as *FL-Tools* are not supported (*19*, *20*). Its local, containerized evaluation *Workflow* and governance by stakeholders for the benchmarking use case supports disclosure control, security and auditing. However, as there is no federated learning channel to use, the technical requirements are not met and RE2 is therefore also not met (*19*, *20*). Datasets are registered with specific labels, giving unharmonized information about them (*21*). Benchmark creators can then harmonize using benchmark-specific data preparers. Therefore, MedPerf lacks a complete semantic and syntactic mapping, earning partial fulfillment of RE1 (*19*, *20*). *Site-Owners* must run the benchmarks, so explorative data discovery as required by RE4 is not available. While MedPerf satisfies several of these requirements, it is engineered specifically for federated evaluation and benchmarking rather than FL. Consequently, it cannot serve as a framework for model training, which is central to the scope of this paper. In summary, MedPerf does not fulfill RE2, RE3, RE4 and RE5 and partially fulfills RE1 (Table 1) and does not support FL.

## NVIDIA FLARE

NVIDIA FLARE is an open-source SDK for FL that facilitates the transition from simulation and research to production-ready, real-world FL deployments (*22*, *23*). Its key contribution is a scalable, enterprise-oriented runtime for multi-partner federated learning Tools. This runtime includes secure provisioning, mutual TLS-based communication, audit logging, role-based management, privacy filtering, and support for various training libraries and *Workflows (22, 23)*. It fully supports RE2 through secure provisioning, authenticated identities, authorization, location policies, data protection filters, secure communication, and logging, as described in its security architecture (*22*, *23*). It partially supports RE3, as jobs and *Workflows* are explicitly deployable objects that can be transferred and reused, but a specific persistence via NVIDIA Flare itself for these jobs does not exist, so versioning is handled externally e.g. via git (*22*, *23*). FL executions are kept partially traceable and reusable through the Job system.
While a Job keeps the track of the FL execution itself including participants via a deploy map and Tools as part of the job, the final results are only accessible via download and not directly a part of the job execution (*22*, *23*) Therefore, RE5 is partially fulfilled. While preprocessing can be implemented in jobs, no specific harmonization capabilities are given, thus partially fulfilling RE1. Considering RE4, dedicated support for federated statistics across *Sites* exists, but semantically the same variables are not automatically harmonized, making the data discovery partially manual and therefore only partially fulfilling RE4 (*22*, *23*). In summary, NVIDIA FLARE partially fulfills RE1, RE3, RE4 and RE5 and fully fulfills RE2 (Table 1).

## OpenFL

OpenFL is an open-source, python based framework for FL that integrates existing machine and deep learning pipelines into collaborative training processes that protect data privacy. This framework focuses on providing a practical framework for TensorFlow- and PyTorch-based *FL-Tools* (*24*)**.** No specific harmonization support is given except from preprocessing capabilities, resulting in no fulfillment of RE1. Considering data discovery, federated analytics allows for manual discovery earning partial fulfillment of RE4. Training plans are manually accepted per dataset, with data being loaded in manually. While security aspects of the FL are incorporated, the manual data selection only gives partial fulfillment of RE2. While FL Plans package *Workflows* including their configuration, no versioning is directly supported and RE3 is therefore only partially met. Similarly, the final model is simply downloaded, losing the trace to the FL plan. Therefore, RE5 is only partially fulfilled (*24*, *25*). In summary, OpenFL does not fulfill RE1, partially fulfills RE2, RE3, RE4 and RE5 (Table 1).

## Personal Health Train

The Personal Health Train (PHT) is a distributed FAIR data paradigm (originated within the GO-FAIR initiative) that shifts the computing model to routing algorithm containers ("trains") across clinical institutions ("stations")(*26*). While PHT promotes FAIR data, the software does not support any specific harmonization or defined data standards, data itself is simply mounted into the trains (*27*). Therefore, RE1 is not met. The PHT architectures natively integrate metadata query endpoints. However, the specification for the datasets is very simple and requires manual discovery over all the stations (*27*). Therefore, RE4 is not met. RE2 is fully met through station-level governance, where local station managers review, authorize, and isolate incoming trains using secure container engines, cryptographic protocols, and local policy enforcement, auditing the whole process. RE3 is fully met because analytical *Workflows* are encapsulated as immutable, version-controlled Docker containers (“trains”) that are registered in central train depots and re-executed reproducibly across *Sites*. However, results of train runs are simply downloaded, losing traceability to their train and therefore only partially meeting RE5 (*26*, *27*). In summary, Personal Health Train does not fulfill RE1 and RE4, partially fulfills RE5 and fully fulfills RE2 and RE3 (Table 1), but FL is sequential (from station to station) and therefore less performant as all other, star shaped architecture based solutions.

## PySyft

PySyft is an open-source library that makes ML easily accessible, secure, and private. This work's main contribution lies in abstracting technical privacy primitives, such as FL, secure multi-party computation, and differential privacy. This makes them easier to use in experimental and development-oriented ML *Workflows*. Therefore, PySyft is particularly relevant as a methodological and programmable foundation for privacy-preserving ML experiments (*28*, *29*). In PySyft, datasets contain mock data to help *Researchers* understand the data structure without disclosing any real data. This allows for data discovery, although this must be done manually, therefore earning partial fulfillment of RE4. RE1 is not fulfilled, data is simply described but not harmonized. However, RE2 is supported because data owners can control access and approve calculations without disclosing raw data. *Workflows* are packaged into code requests, containing all code requested to be run. A code request is therefore a reusable, immutable representation of the *Workflow* (*28*, *29*). This results in partial fulfillment of RE3, as versioning is not included in this design. Lastly, results of an FL execution are simply retrieved, losing their connection to the code request, only earning partial fulfillment of RE5 (*28*, *29*). In summary, PySyft does not fulfill RE1, partially fulfills RE3, RE4 and RE5 and fully supports RE2 (Table 1).

## Rhino FCP

As a commercial provider, Rhino FCP combines federated computing, distributed pipelines, and a harmonized data layer to enable secure data collaboration. Its main contribution is the operationalization of federated learning as a production-ready, multi-partner framework. However, as a commercial solution, individual consortia are clearly separated, and the purpose of Rhino FCP is only the support of a single project in their FL (*30*). In this regard, RE1 is only partially met, since while Rhino enables data harmonization and AI-assisted mapping, these systems are per project only and no cross project support, e.g. based on ontologies, exists (*30*). RE2 is enabled by role-based access control, secure containers, privacy-enhancing techniques, encryption, and comprehensive audit logging (*30*). RE3 and RE5 are partially met because, although federated MLOps and AI lifecycle management are mentioned, the reviewed public pages do not specify immutable *Workflow* versions or a complete model for tracing the provenance of projects, cohorts, approvals, and artifacts (*30*). RE4 is partially met, as federated analytics and the exploration of protected data are offered, but this is, similarly to the harmonisation, only per project without cross domain support as possible via ontologies (*30*). In summary, Rhino FCP partially fulfills RE1, RE3, RE4 and RE5 and fully fulfills RE2 (Table 1), but it should be noted that it's a commercial, non open source solution focused on single projects.

## Vantage6

Vantage6 is an open-source framework designed for privacy-preserving FL and secure data analysis from distributed sources (*31*). Its main contribution is its flexible, container-based architecture, which supports both horizontal and vertical data silos. This architecture grants participating institutions a high degree of autonomy over their data, infrastructure, and executable algorithms. Thus, Vantage6 specifically addresses the organizational and technical heterogeneity of real-world, multicenter research networks (*31*). RE1 is not met, there is no common system for semantic and syntactic harmonization at the local level, the system of using docker images as processing *Tools* may be used to perform preprocessing for harmonization, but the framework has no specific support (*31*). This architecture and the data center operator's control and log access over executable algorithms support RE2 at the *Platform* level (*31*). Federated Access is gained via notebook style access. This only partially fulfills RE3 as versioning isn't incorporated, but RE5 is fully met as the notebook contains the complete trace and outputs of an FL execution (*31*). RE4 is partially met: Distributed aggregation algorithms can be used to calculate feasibility information, and local policies can restrict the results; however, no cross domain discovery mechanism exists, *Sites* must be individually queried and manually analyzed and mapping between *Sites* must be done manually without ontology annotation support available by the system (*31*). In summary, Vantage6 does not fulfill RE1, partially fulfills RE3 and RE4 and fully fulfills RE2 and RE5 (Table 1).

# S6: Glossary

| Term | Meaning |
| --- | --- |
| Auditor | The responsible person, which is allowed to audit *Tools* |
| Building-Pipeline | Builds securely all Docker containers for each *Tool*, also checks for voluntary, unit tests and malware |
| Connector | The Connector describes the configuration of an whole ETL pipeline towards a specific *Schema*. Created on specific input but can be reused with new input |
| Data discovery query (DDQ) | A federated query is fired to all connected *Sites* to aggregate in a private, preserving manner how many patients fit to the query criteria (rounded), explained in S2 |
| FL-Net Platform | The coordination instance of FL-Net to which all FL-Net *Sites* connect to. Also simply called *Platform* |
| FL-Net Site | A connected clinic to the FL-Net network, also simply called *Site* |
| FL-Request | The request for FL created by a *Researcher* and distributed to the *Sites*. Is linked to a *Project*. Can request the model of the FL to be published or kept private at each *Site* |
| $N_{DT}$ | Any specific DataType Node. A DataType Node represents the syntactic information of a variable |
| $N_O$ | Any specific node in the ontology DAG. An Ontology Node represents the semantic information of a variable |
| $N_S$ | Any specific *Schema* Node. A *Schema* Node represents a linkage of a DataType and Ontology Node |
| $P_{O\text{-}DT}$ | A pair of an $N_O$ and $N_{DT}$ describing a variable |
| Project | The entity describing the whole FL process in FL-Net. Contains a connected and ran *DDQ*, a selection of variables based on *Schemas* that can fulfill the *DDQ*, a *Workflow* and meta information such as a name and description |
| Researcher | A *Platform* user who launches *DDQ*, creates a FL *Project* from it, and carries it out accordingly |
| Schema | The tree of $N_S$ describing a data standard |
| Site-Owner | The responsible person for managing data inside a *Site* |
| Tool | A *Tool* describes code that can be used for ETL, FL, or inference (*5*). Therefore, *ETL-Tools*, *FL-Tools* and *Inference-Tools* exist |
| Tool-Store | The *Platforms Tool-Store* contains all *Tools* used in the FL-Net |
| Workflow | A *Workflow* describes the execution order of Tools as an DAG (*5*) |

# Supplementary References